\documentclass[conference]{IEEEtran}

\def\IEEEtitletopspace{18pt}
\usepackage[left=54pt, right=54pt, bottom=54pt, top=54pt]{geometry}

\IEEEoverridecommandlockouts
\usepackage{cite}
\usepackage{amssymb,amsfonts}
\usepackage{times}
\usepackage{algorithmic}
\usepackage{graphicx}
\usepackage{textcomp}
\usepackage{pifont}
\usepackage{float} %
\usepackage{hyperref}
\usepackage{xcolor}
\usepackage{multirow}     %

\def\BibTeX{{\rm B\kern-.05em{\sc i\kern-.025em b}\kern-.08em
    T\kern-.1667em\lower.7ex\hbox{E}\kern-.125emX}}
    
\begin{document}

\newcommand{\dsname}{RealDriveSim}

\title{\dsname{}: A Realistic Multi-Modal Multi-Task Synthetic
Dataset for Autonomous Driving\\
}

\author{
    Arpit Jadon\textsuperscript{1,3*},
    Haoran Wang\textsuperscript{3},
    Phillip Thomas\textsuperscript{4},
    Michael Stanley\textsuperscript{4},
    S. Nathaniel Cibik\textsuperscript{4},\\
    Rachel Laurat\textsuperscript{4},
    Omar Maher\textsuperscript{4},
    Lukas Hoyer\textsuperscript{5},
    Ozan Unal\textsuperscript{2,5*},
    Dengxin Dai\textsuperscript{2,5}\\[0.40ex]
    \textsuperscript{1}German Aerospace Center, Braunschweig, Germany\\
    \textsuperscript{2}Computer Vision Lab, Huawei Research Center Zurich, Switzerland\\
    \textsuperscript{3}Max Planck Institute for Informatics, Saarland Informatics Campus, Germany\\
    \textsuperscript{4}Parallel Domain, San Francisco, CA, USA\\
    \textsuperscript{5}ETH Zurich, Switzerland\\[0.3ex]
    \texttt{arpit.jadon@dlr.de, hawang@mpi-inf.mpg.de, phillipthomas@live.de}\\
    \texttt{\{michaelhamerstanley, nate.cibik\}@gmail.com, rachel.laurat@paralleldomain.com}\\
    \texttt{omar@monta.ai, \{ozan.unal, dengxin.dai\}@huawei.com, lhoyer@vision.ee.ethz.ch}\\
    \thanks{\textsuperscript{*}equal contribution.}

}

\maketitle

\begin{abstract}

As perception models continue to develop, the need for large-scale datasets increases. However, data annotation remains far too expensive to effectively scale and meet the demand. Synthetic datasets provide a solution to boost model performance with substantially reduced costs. However, current synthetic datasets remain limited in their scope, realism, and are designed for specific tasks and applications. In this work, we present \dsname, a realistic multi-modal synthetic dataset for autonomous driving that not only supports popular 2D computer vision applications but also their LiDAR counterparts, providing fine-grained annotations for up to 64 classes. We extensively evaluate our dataset for a wide range of applications and domains, demonstrating state-of-the-art results compared to existing synthetic benchmarks. The dataset is publicly available at https://realdrivesim.github.io/.

\end{abstract}

\begin{IEEEkeywords}
Semantic Scene Understanding, LiDAR, Object Detection, Segmentation, Autonomous Driving
\end{IEEEkeywords}

\section{Introduction}

With the ever-increasing popularity of autonomous driving, research and development of visual perception models have gained further traction. This has resulted in the rise of highly engineered solutions that rely on multi-sensor and multi-modality training data to address multiple vision tasks~\cite{zhang2021survey, unal2021improving, bhattacharjee2022mult, li2022deepfusion}.

\begin{figure}[htbp]
    \centering
    \includegraphics[width=0.797\columnwidth]{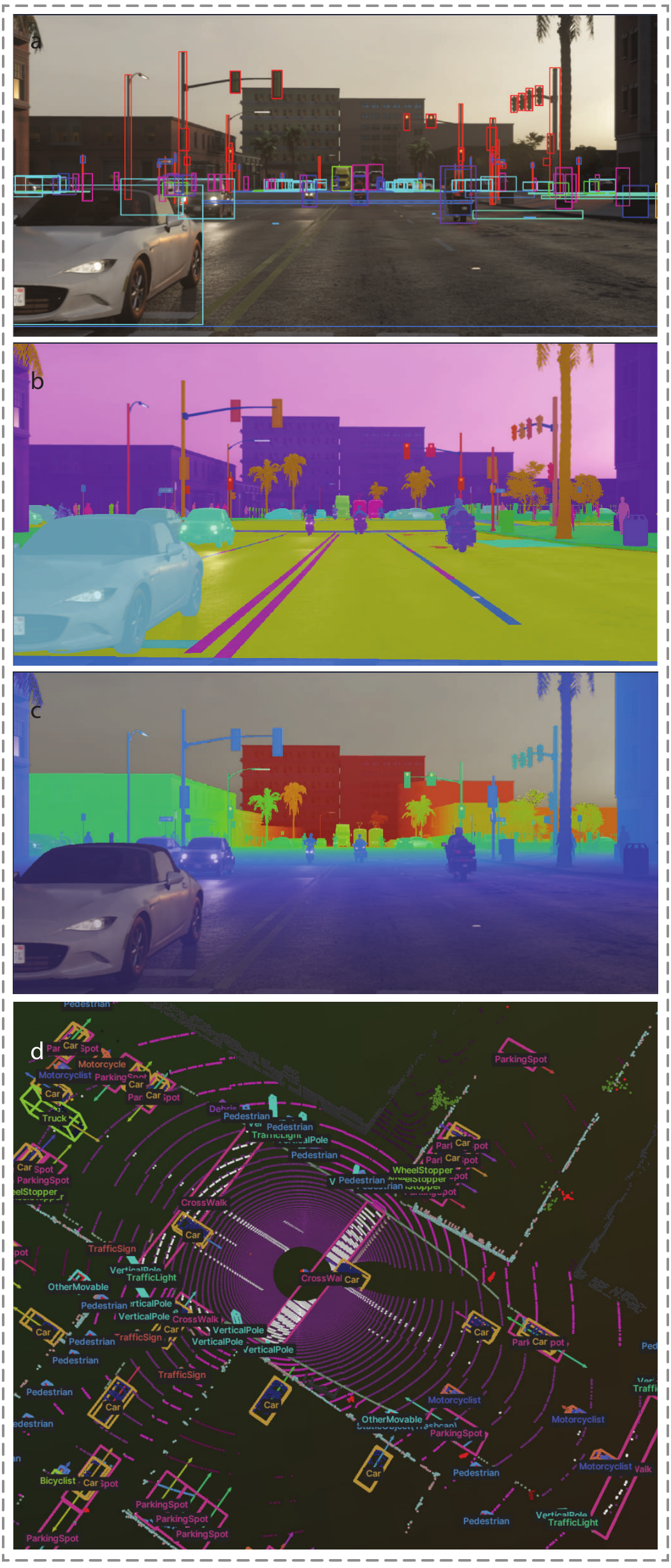}
    \caption{Example from the \dsname{} dataset showcasing a) 2D object detection; b) 2D semantic segmentation; c) depth estimation; and d) 3D semantic segmentation and object detection annotations on the camera and HDL-64 Velodyne sensors.}
    \label{fig:teaser}
\vspace{-10px} \end{figure}

A major bottleneck for real-world deployability of such deep learning models remains the necessity of large-scale driving datasets that can not only cover a plethora of varying scenarios but also support a wide range of tasks and modalities. However, collecting and annotating such datasets remains far too costly to easily and effectively scale up. To this end, many recent efforts have focused on developing data-efficient methods under the umbrella of \textit{weakly-} or \textit{semi-}supervised settings~\cite{unal2022scribble,xu2020weakly,zhang2021weakly, unal2023discwise, wang2022semi, kwon2022semi}, with the unified goal of reducing annotation costs while retaining an adequate level of performance. However, even under such data-efficient structures, given the diversity of available sensors (e.g. camera, LiDAR) and computer vision problems (e.g. semantic segmentation, object detection, depth estimation), labeling large-scale data for each setup and application remains impractical. For certain tasks, such as optical flow or occluded object tracking, correct and granular annotation is nearly impossible regardless of time and cost.

Consequently, with the continually developing scene of computer graphics and scene rendering in the last decade, there has been an increased interest in generating and exploiting synthetic driving data to improve data variability and introduce additional complex scenes in a controlled manner with negligible monetary costs. The game Grand Theft Auto V (GTA V) has been frequently used to generate synthetic data with automatic annotations~\cite{richter2016playing_GTA, Richter_2017_ICCV_Viper, hurl2019precise_presil}. CARLA~\cite{Dosovitskiy17_CARLA_Sim} is another open source simulator that has been used extensively to generate various synthetic driving datasets~\cite{sun2022shift, alberti2020idda, sekkat2022synwoodscape, sekkat2024amodalsynthdrive}. 

Real-world driving scenarios can be extremely diverse, ranging from multiple cities~\cite{Cordts2016Cityscapes}, different countries~\cite{varma2019idd,yu2020bdd100k} to different weather and daytime conditions~\cite{sakaridis2024acdcadverseconditionsdataset}. Thus, a good synthetic dataset must not only strive for realism but also encompass a diverse range of possible scenarios. Moreover, to serve as a new benchmark in the computer vision community for evaluating various methods on real-world datasets, a synthetic dataset must align with existing real datasets in terms of the granularity of its semantic classes. Semantic classes used in Cityscapes~\cite{Cordts2016Cityscapes} have been widely adopted in various benchmark real datasets~\cite{varma2019idd, sakaridis2024acdcadverseconditionsdataset, yu2020bdd100k}. Most methods~\cite{vu2019advent, hoyer2022daformer, loiseau2024reliability} within the synthetic-to-real-domain adaptation and transfer learning space use the same set of classes for performance comparison. Therefore, having at-least this level of semantic class granularity in a synthetic dataset is essential for its convenient adoption within the vision community. Furthermore, a multi-modal sensor setup, such as camera and LiDAR along with a comprehensive annotation suite that supports various tasks such as semantic and panoptic segmentation, object detection, depth estimation, optical flow, multi-object tracking, and scene flow, helps in building a robust perception system for autonomous driving. Most existing synthetic datasets fail to strike a balance among the aforementioned qualities. Consequently, despite years of advancements in computer graphics and scene rendering, GTA~\cite{richter2016playing_GTA} and Synthia~\cite{ros2016synthia_synthia_rand_seqs} continue to be the most commonly used synthetic datasets for autonomous driving applications.

\begin{figure*}[htbp]
    \centering

    \begin{minipage}{0.33\textwidth}
        \centering
        \includegraphics[width=\textwidth]{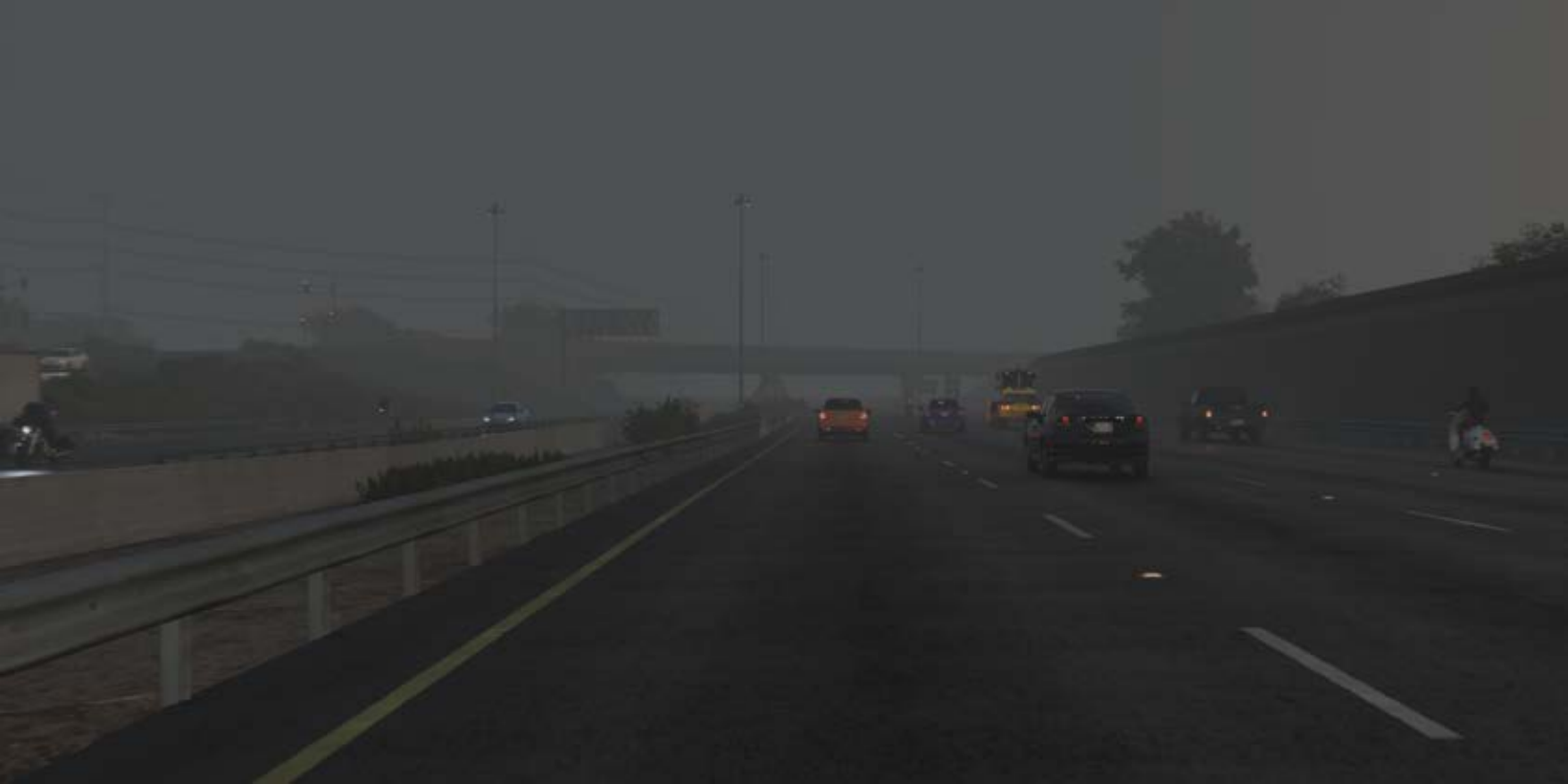} %
    \end{minipage}
    \hspace{-1.95mm}
    \begin{minipage}{0.33\textwidth}
        \centering
        \includegraphics[width=\textwidth]{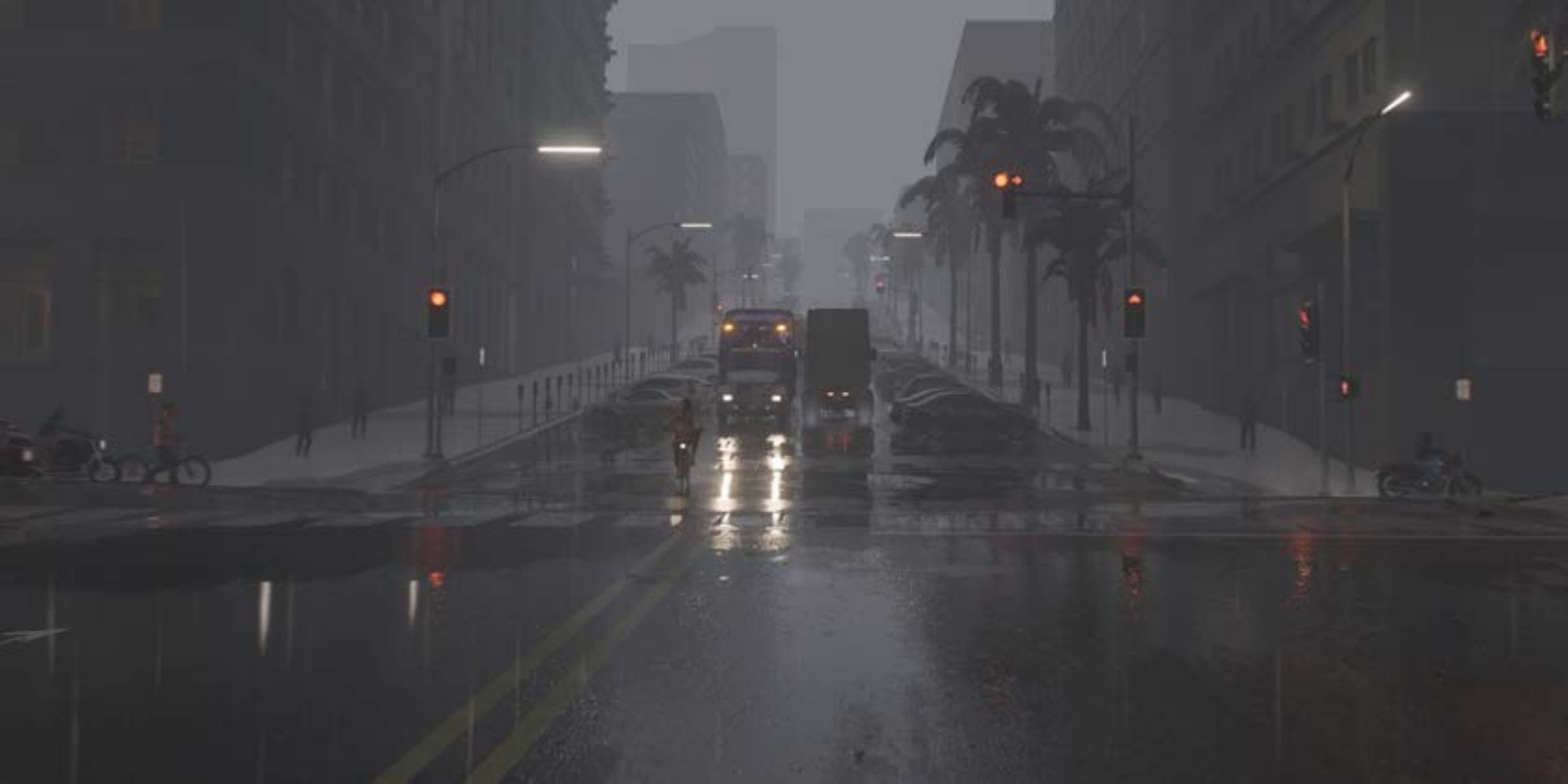} %
    \end{minipage}
    \hspace{-1.95mm}
    \begin{minipage}{0.33\textwidth}
        \centering
        \includegraphics[width=\textwidth]{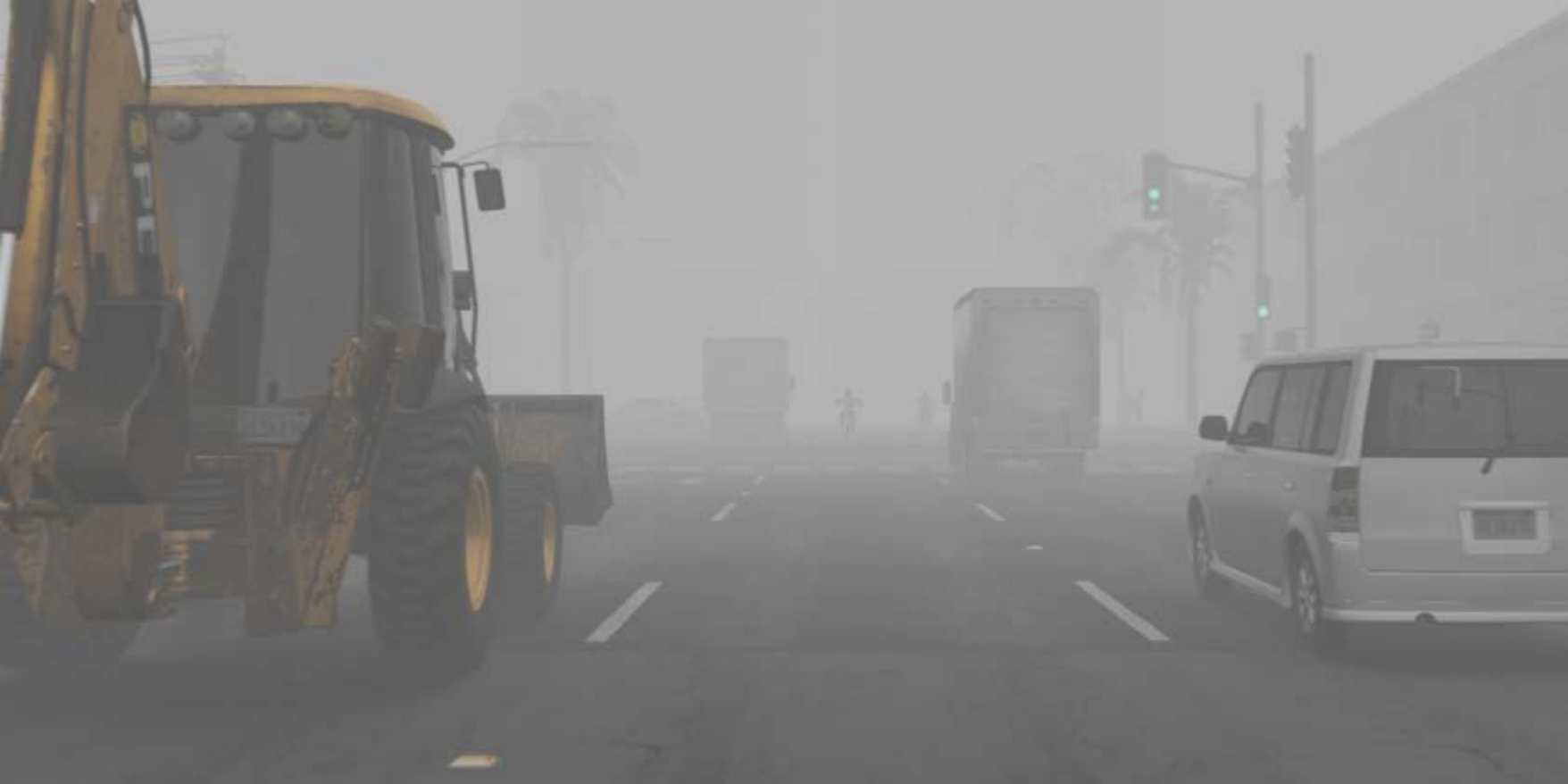} %
    \end{minipage}
    
    \vspace{0.45mm} %

    \begin{minipage}{0.33\textwidth}
        \centering
        \includegraphics[width=\textwidth]{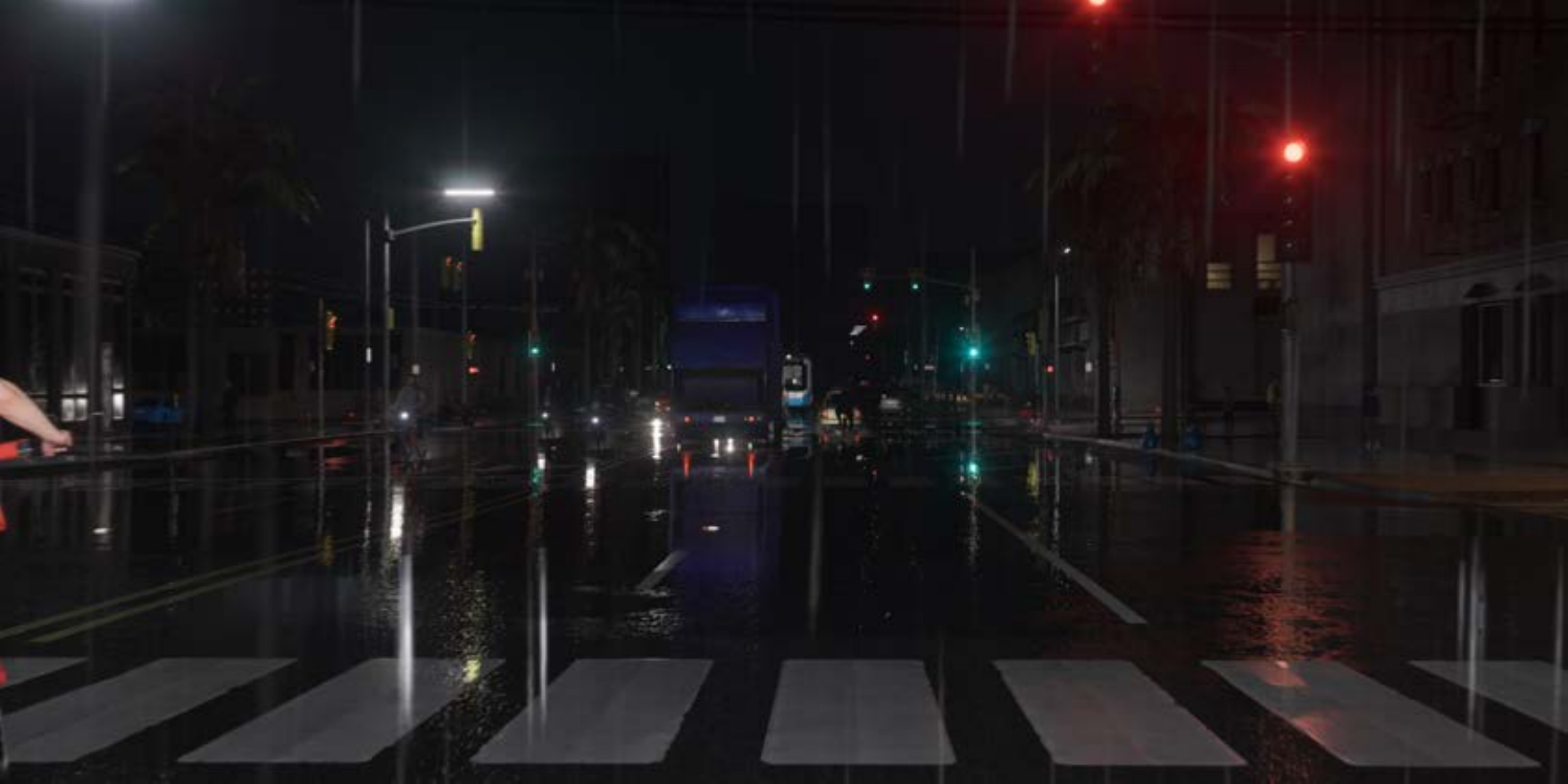} %
    \end{minipage}
    \hspace{-1.95mm}
    \begin{minipage}{0.33\textwidth}
        \centering
        \includegraphics[width=\textwidth]{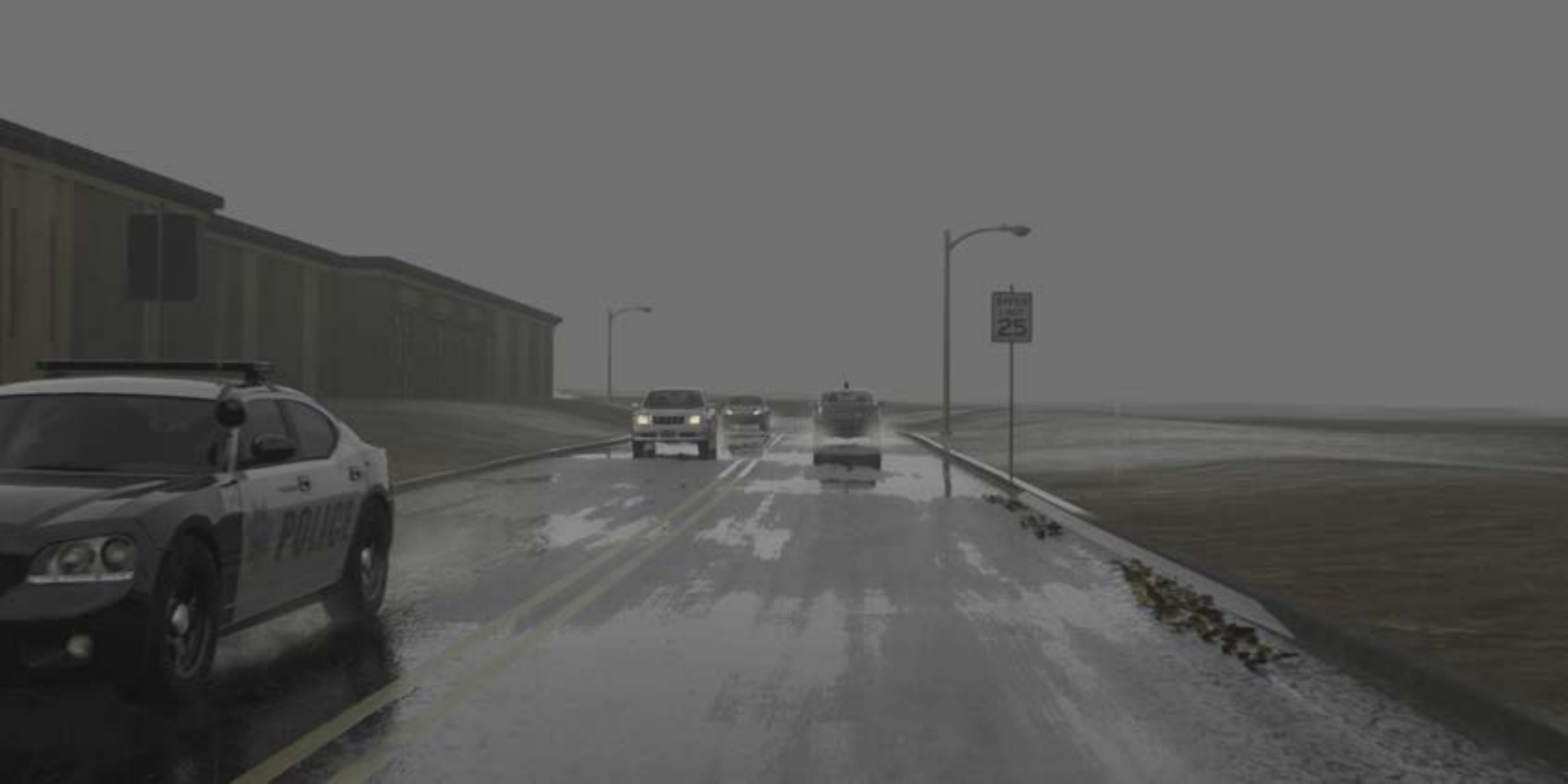} %
    \end{minipage}
    \hspace{-1.95mm}
    \begin{minipage}{0.33\textwidth}
        \centering
        \includegraphics[width=\textwidth]{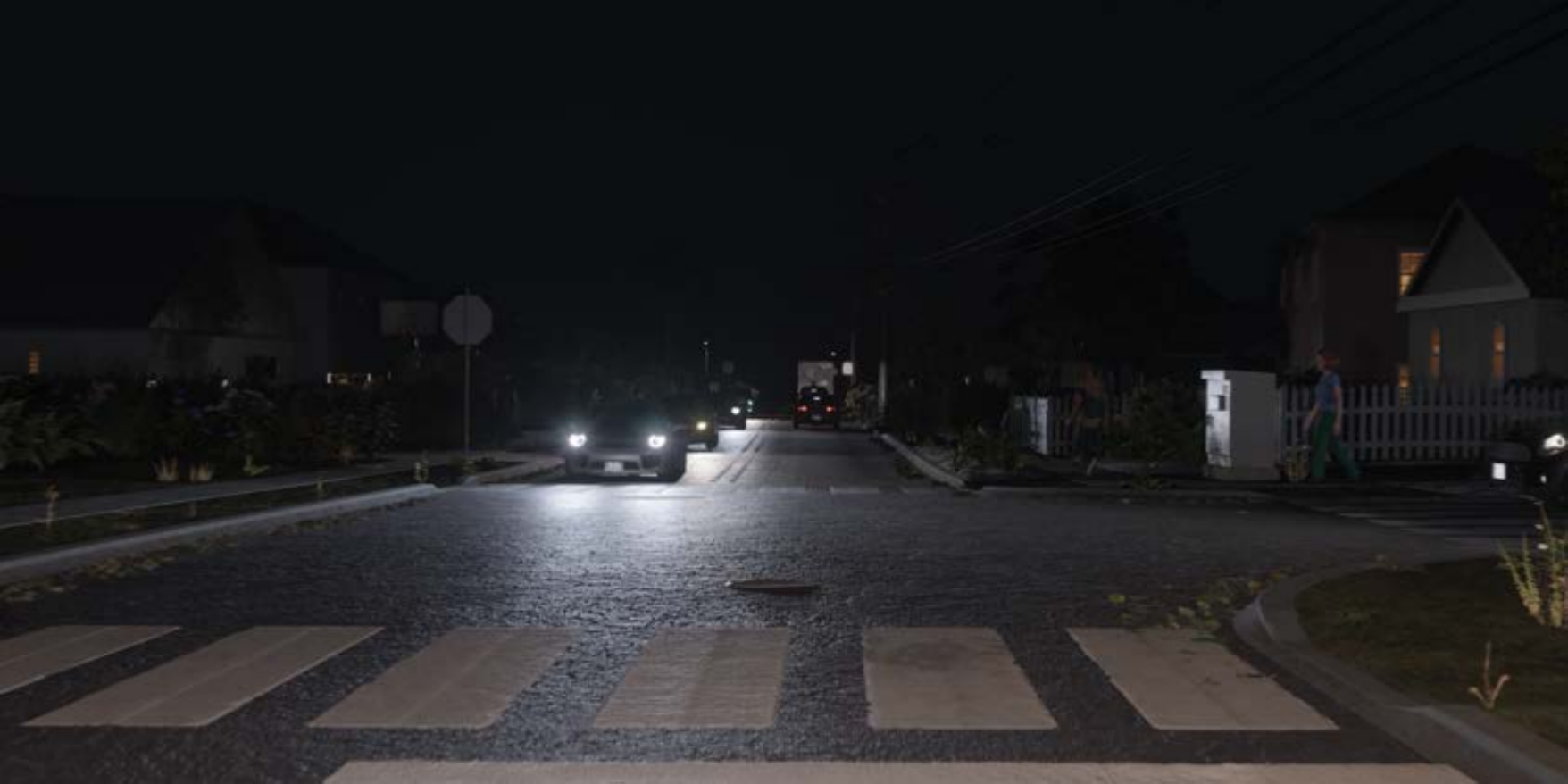} %
    \end{minipage}

    \vspace{0.55mm} %

    \begin{minipage}{0.247\textwidth}
        \centering
        \includegraphics[width=\textwidth]{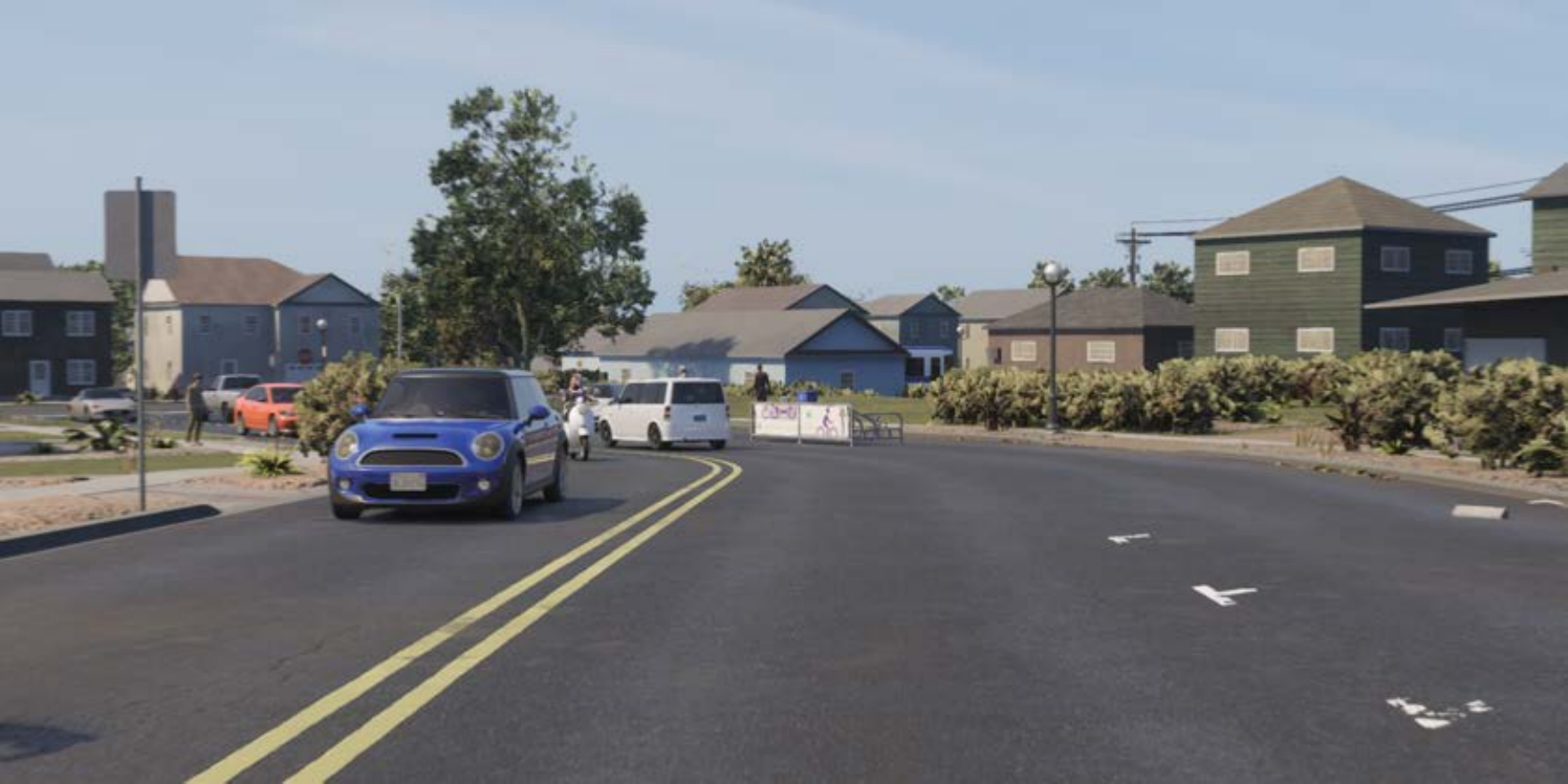} %
    \end{minipage}
    \hspace{-1.95mm}
    \begin{minipage}{0.247\textwidth}
        \centering
        \includegraphics[width=\textwidth]{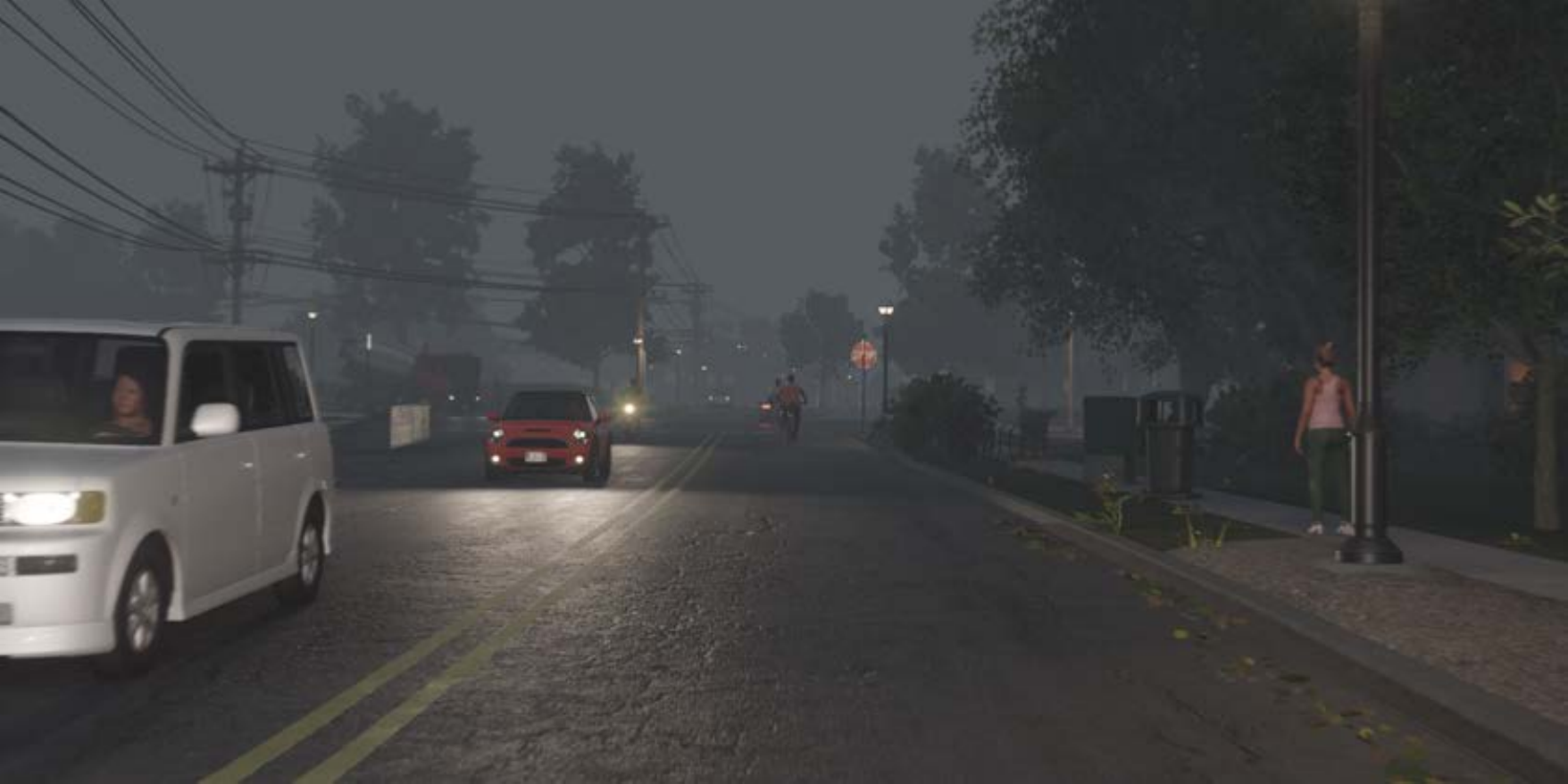} %
    \end{minipage}
    \hspace{-1.93mm}
    \begin{minipage}{0.247\textwidth}
        \centering
        \includegraphics[width=\textwidth]{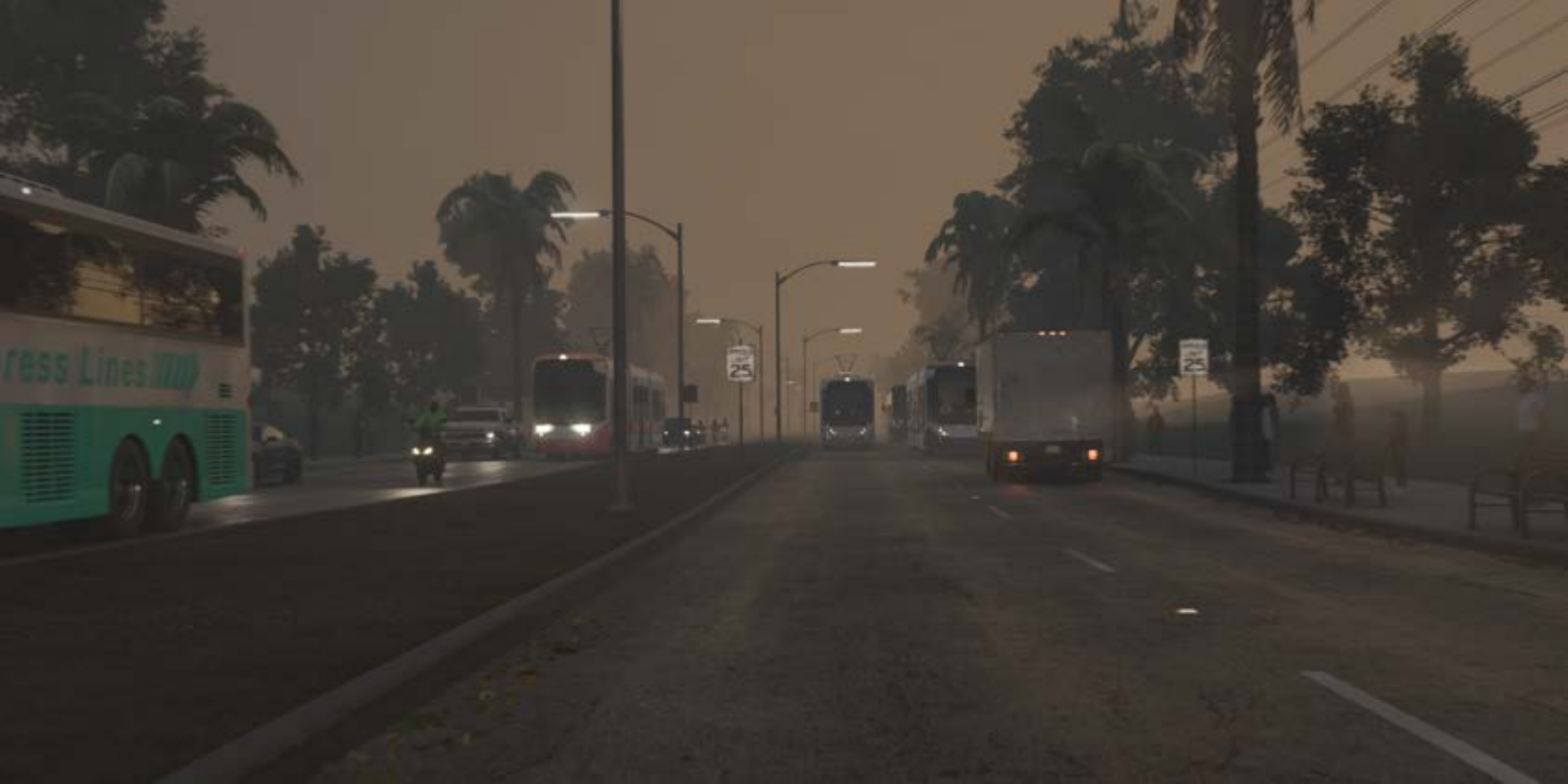} %
    \end{minipage}
    \hspace{-1.95mm}
    \begin{minipage}{0.247\textwidth}
        \centering
        \includegraphics[width=\textwidth]{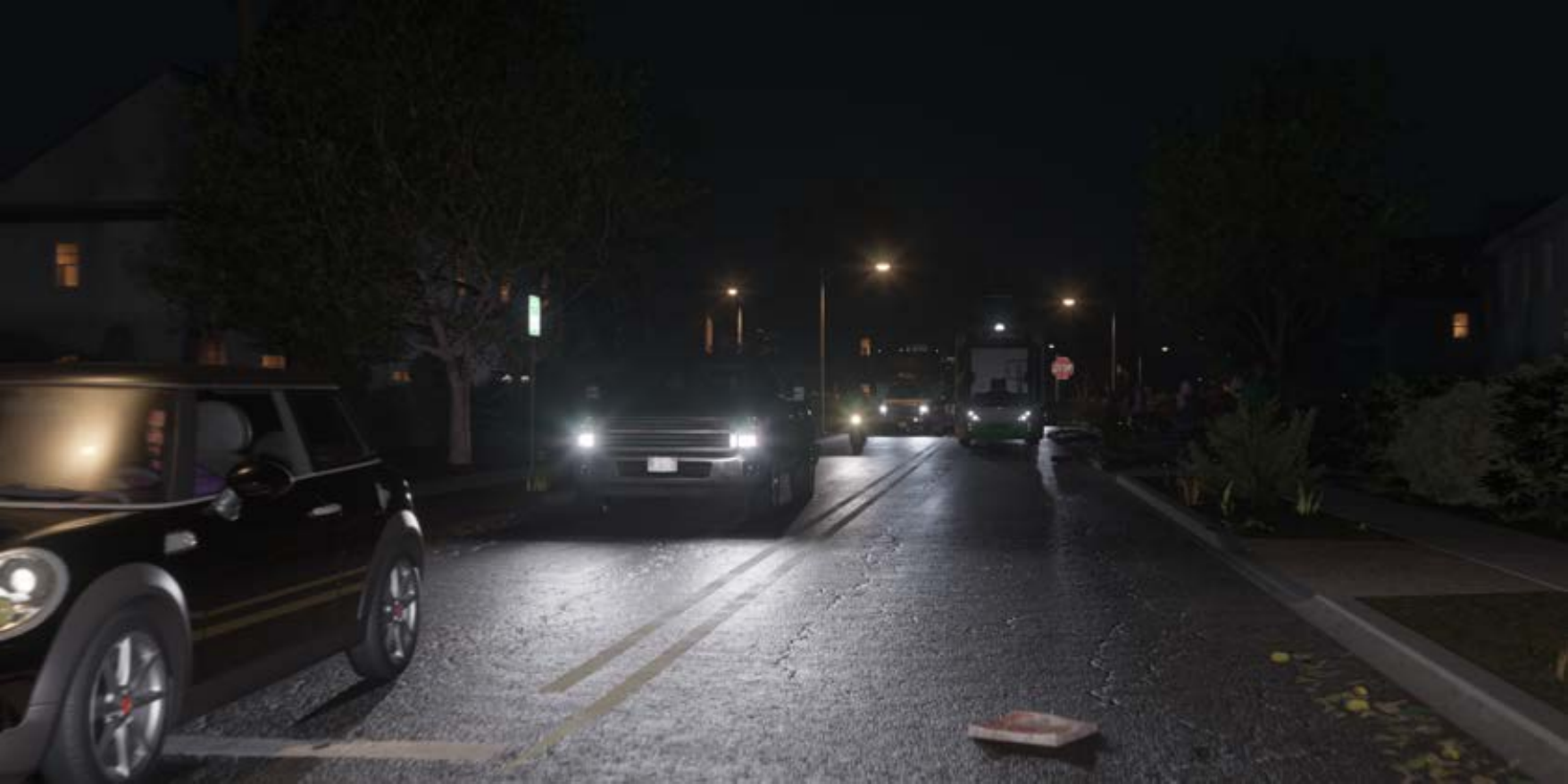} %
    \end{minipage}

    \caption{Example samples from \dsname{}: first row shows different adverse weather conditions - cloudy, rainy, and foggy weather respectively. Second row consists of images with mixed adverse conditions (rain+night, fog+rain, and fog+night). The last row depicts different daytime conditions: sunny, dawn, dusk, and night respectively.}
    \label{fig:bigpdv4_example_samples}
\vspace{-10px} \end{figure*}

\begin{figure*}[htbp]
    \centering
    \includegraphics[width=\textwidth, height=4.76cm]{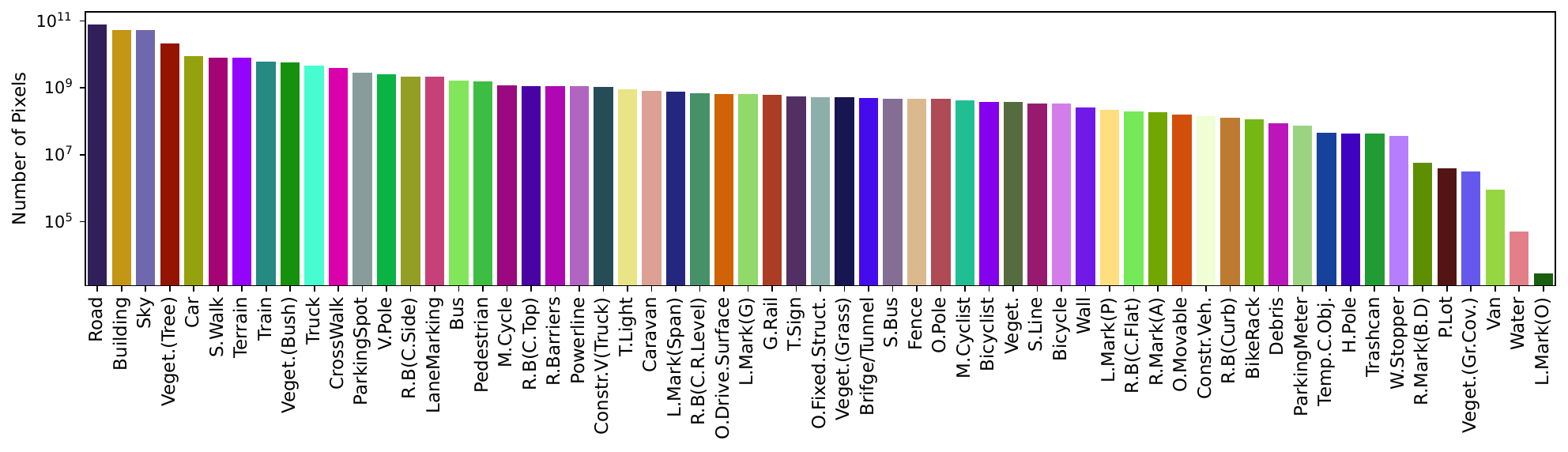}
    \caption{Pixel-wise semantic class statistics for images from \dsname{}. Y-axis is scaled logarithmically. Note that the LiDAR counterpart contains three extra classes. Abbreviations: O. (other), L. (lane), R. (road), R.B (road boundary), G. (guard), H. (horizontal), and V. (vertical). }
    \label{fig:bigpd_class_stats}
\vspace{-10px}
\end{figure*}

In this work, addressing the above-mentioned issues, we introduce \dsname{}, a multi-modal multi-task synthetic dataset for autonomous driving with high realism and diversity. The dataset consists of 6,689 scenes or short sequences corresponding to 133,780 frames generated across 10 maps of urban, suburban, and highway environments. \dsname{} provides sensor data that includes images captured from a front-facing camera and 3D point clouds obtained from 32-beam LiDAR and 64-beam LiDAR. The annotation suite supports a range of perception tasks including 2D/3D object detection, 2D/3D semantic segmentation, 2D/3D instance segmentation, depth estimation, 2D optical flow estimation, and 3D scene flow estimation. Certain annotation modalities from our dataset are presented in Fig.~\ref{fig:teaser} whereas Fig.~\ref{fig:bigpdv4_example_samples} offers a glimpse of the dataset's high realism with images captured under different weather and daytime conditions. To the best of our knowledge, \dsname{} is the only high-fidelity synthetic autonomous driving dataset to support a wide range of 2D perception tasks while also providing corresponding LiDAR data and their point-wise ground truth annotation for various 3D perception tasks. Concurrently, \dsname{} demonstrates a higher level of realism than existing synthetic datasets. This opens the door to not only investigating the use of synthetic datasets for multi-modality but also multi-task approaches, both in the 2D and 3D perception domains. Across all modalities, our dataset provides fine-grained class definitions for up to 64 initial classes (Fig.~\ref{fig:bigpd_class_stats}) that can be further mapped to fit any popular dataset, allowing joint training, transfer learning, or domain adaptation with any real-world dataset without any compromise.

We extensively evaluate \dsname{} using real-world datasets and compare it to existing works. Specifically for 3D semantic segmentation, \dsname{} outperforms SynLiDAR~\cite{xiao2022synlidar} on both SemanticKITTI~\cite{behley2019semantickitti} and SemanticPOSS~\cite{pan2020semanticposs} in the transfer learning, joint training and varying degrees of fine-tuning setting. On similar tasks, \dsname{} outperforms PreSIL~\cite{hurl2019precise_presil} on KITTI~\cite{geiger2012we} for 3D object detection. For 2D object detection, \dsname{} outperforms existing synthetic datasets in almost all source-only and joint training settings on Cityscapes~\cite{Cordts2016Cityscapes} and BDD100k~\cite{yu2020bdd100k}. For 2D semantic segmentation, \dsname{} achieves comparable results to existing datasets in a source-only setup on Cityscapes and BDD100k. Again, we would like to draw the reader's attention to the wide usability of the \dsname{} dataset that accounts for all tasks thanks to its multi-sensor support and fine-grained class definitions. To be able to compare our dataset with existing benchmarks, we use multiple synthetic datasets tailored for a significant range of tasks or class granularity.

\noindent Our contributions can be summarised as follows:
\begin{enumerate}
    \item We construct the \dsname{} dataset which provides multi-modal data with both camera and LiDAR sensor outputs while demonstrating a high level of realism.
    \item We provide a comprehensive annotation suite supporting a wide range of 2D and 3D perception tasks with fine-grained class definitions of up to 64 semantic classes, allowing full mapping to all popular real-world datasets. 
    \item We extensively evaluate the proposed dataset and show that it achieves state-of-the-art performance on a multitude of tasks, modalities, and real-world datasets. Creating a strong case for \dsname{} being adopted as a new benchmark synthetic dataset for various driving related vision tasks.
\end{enumerate}

\begin{table*}[htbp]
    \centering
    \tabcolsep=0.060cm
\caption{Summary of the supported tasks from various commonly used synthetic datasets. \dsname{} not only supports a wide range of applications in the camera domain, but also provides rich annotations for popular LiDAR perception tasks. Adverse-W, MOT, and Sem-Cls. refer to adverse weather, multi-object tracking, and the number of semantic classes in the dataset respectively.} 

\begin{tabular}{|l|cccccccc|cccccc|} 
 \cline{2-15}
 \multicolumn{1}{c|}{} & \multicolumn{8}{c|}{Camera} & \multicolumn{6}{c|}{LiDAR} \\
 \hline
 Dataset & Adverse-W & 2D Seg. & Sem-Cls. & 2D Det. &3D Det. & Depth & Optical-Flow & MOT & 3D Det. & 3D Seg. & Sem-Cls. & Scene-Flow & MOT & SLAM \\ 
 \hline
 SYNTHIA \cite{ros2016synthia_synthia_rand_seqs} & \ding{51} & \ding{51} & $22$ & \ding{51} & \ding{51}
 & \ding{51} & \ding{55} & \ding{55} & -  & - & - & - & - & -  \\

 GTA-V \cite{richter2016playing_GTA} & \ding{51} & \ding{51} & $19$ & \ding{55}  & \ding{55} & \ding{55}  & \ding{55} & \ding{55} & - & - & - & - & - & -  \\
 VIPER \cite{Richter_2017_ICCV_Viper} & \ding{51} & \ding{51} & $32$ & \ding{51} & \ding{51} & \ding{55} & \ding{51} & \ding{51} & - & - & - & - & - & -  \\
 Synscapes \cite{wrenninge2018synscapes} & \ding{55} & \ding{51} & $19$ & \ding{51} & \ding{51} &  \ding{51} & \ding{55} & \ding{55} & - & - & - & - & - & - \\
 SHIFT \cite{sun2022shift} & \ding{51} & \ding{51} & $23$ & \ding{51} & \ding{51} & \ding{51} & \ding{51} & \ding{51} & \ding{51} & \ding{55} & - & \ding{55} & \ding{51} & \ding{55} \\
 PreSIL \cite{hurl2019precise_presil} & \ding{55} & \ding{51} & $12$ & \ding{51} & \ding{51} & \ding{51} & \ding{55}   & \ding{55} & \ding{51} & \ding{55} & $12$ & \ding{55} & \ding{55} & \ding{55} \\
 SynLIDAR \cite{xiao2022synlidar} & - & - & - & -  & - & -  & -  & - & \ding{55} & \ding{51} & $32$ & \ding{55} & \ding{55} &  \ding{55}\\
 \hline
 \dsname{} (Ours) & \ding{51} &\ding{51} & $61$ & \ding{51} &\ding{51} &\ding{51} & \ding{51} & \ding{51} & \ding{51} & \ding{51} & $64$ & \ding{51} & \ding{51} & \ding{51}  \\
 \hline
\end{tabular}
\label{tab:task_comp}
\vspace{-10px} \end{table*}

\section{Related Work}

Synthetic data generation and utilization has been a core research topic in computer vision, especially for the autonomous driving community where interesting and out-of-distribution data is scarce and difficult to obtain.

One of the initial works to introduce such a synthetic driving dataset was SYNTHIA~\cite{ros2016synthia_synthia_rand_seqs}, in which authors demonstrated that synthetic pretraining can bring significant performance benefits when testing on real-world driving datasets. In SYNTHIA, a virtual world is simulated using Unity to generate realistic synthetic images to primarily aid the task of semantic segmentation. Different cities are constructed by placing fundamental assets like traffic signs, lamp poles and people. In the context of modern computer graphics and compute power, the photorealism and diversity of SYNTHIA remain highly limited.

While SYNTHIA showed the effectiveness of synthetic datasets for real-world performance and ease of scene rendering with precise ground truth annotations, it had a big realism and diversity gap compared to the actual road scenes. Moreover, the generation and construction of virtual worlds that suit the specific needs of researchers still required considerable effort and expertise. To further reduce this need, authors in~\cite{richter2016playing_GTA} utilized a commercially available video game Grand Theft Auto V (GTA-V) to render a synthetic dataset. By using a wrapper between the game engine and the operating system to record and alter the communication between the engine and the graphics hardware, they captured 25k annotated images within a 49 hours period. The dataset contains all 19 semantic evaluation classes compatible with Cityscapes~\cite{Cordts2016Cityscapes} and has been the epicenter of many synthetic-to-real perception tasks for 2D semantic segmentation. Following GTA-V~\cite{richter2016playing_GTA}, VIPER~\cite{Richter_2017_ICCV_Viper} extends the dataset with high-resolution video sequences consisting of 254k frames, commonly sampled at $1/10$ for training. However, since these datasets were collected from a video game in 2016, there exists an obvious lack of realism compared to real driving scenes.

Synscapes~\cite{wrenninge2018synscapes} is a photorealistic synthetic driving dataset designed to be similar to Cityscapes~\cite{Cordts2016Cityscapes} in structure and content. It contains 25k images rendered with panoptic segmentation masks, 2D bounding boxes, 3D bounding boxes, and depth maps. Unlike GTA-V, Synscapes does not capture data by simulating ego vehicle movement but follows a procedural approach for each individual scene. While this greatly increases diversity in the dataset, it also prohibits many applications that require temporal information.

In the recent years, there has seen an influx of many synthetic datasets created using the CARLA simulator~\cite{Dosovitskiy17_CARLA_Sim}. One such commonly used dataset is SHIFT~\cite{sun2022shift}, a multi-task synthetic driving dataset. By employing CARLA, authors take advantage of the built-in support for multiple sensors (with the exception of the SPAD-LiDAR), preexisting virtual maps and the ability to simulate agent behaviour. 
The focus of SHIFT is to study domain variances in autonomous driving scenarios by leveraging continuous domain shifts in the data. Therefore, a portion of the dataset is built on scenes where a static sensor setup renders a sequence under a continuous domain shift (e.g. from sunny to rain or from day to night). There are other CARLA based synthetic datasets as well. For instance, IDDA~\cite{alberti2020idda} supports semantic segmentation and depth estimation while focusing on domain adaptation and generalization for semantic segmentation. AmodalSynthDrive~\cite{sekkat2024amodalsynthdrive} is a multi-modal multi-task dataset that targets the problem of object recognition under occlusions. Although CARLA enables the collection of diverse scenes and annotation modalities, various limitations restrict the widespread adoption of its derived datasets. Despite recent progress in photorealistic data generation, CARLA significantly struggles in replicating realistic driving scenes, as it is primarily designed for development and testing of autonomous driving systems. Since CARLA is not explicitly optimized for generating highly realistic data, derived datasets come with high domain gap and limited realism resulting in artificial looking scenes that lack various details. Moreover, these datasets lack the required semantic class granularity. Most CARLA datasets do not even have all the 19 semantic classes used for evaluation in Cityscapes, ACDC, BDD, and various other real driving datasets. This highlights the incompatibility of CARLA based datasets with existing real datasets making it even more difficult to readily adopt such synthetic datasets for various studies. SHIFT, IDDA, and AmodalSynthDrive contain 23, 24, and 21 unique unique semantic classes respectively. Whereas, in the same order, these datasets contain only 12, 15, and 17 overlapping classes respectively w.r.t to the 19 commonly used Cityscapes evaluation classes.  

While many datasets have investigated synthetic data generation and utilization for 2D perception tasks, the inclusion of a LiDAR sensor remains limited in the literature. Following GTA-V, PreSIL~\cite{hurl2019precise_presil} extends the wrapper built for the video game to also render 3D point clouds from a simulated LiDAR sensor. They provide 3D bounding box annotations for an additional 50k frames. SynLiDAR~\cite{xiao2022synlidar} is the only synthetic dataset that supports LiDAR semantic segmentation, providing 20k scans with 32 semantic classes that can be mapped to popular real-world datasets such as SemanticKITTI~\cite{behley2019semantickitti} and SemanticPOSS~\cite{pan2020semanticposs}. Compared to SynLiDAR, \dsname{} provides a higher granularity of semantic class definitions that cover a wider range of real-world datasets. \dsname{} also provides ground truth annotations for a large number of other 2D and 3D perception tasks. 

Tab.~\ref{tab:task_comp} provides an overview of popular synthetic datasets and their currently supported tasks. As evident, the current literature still lacks an all-purpose synthetic dataset that is not only diverse and annotation-rich with fine-grained class granularity but also photorealistic, reflecting advancements in computer graphics and scene rendering. 

\begin{figure}[htbp]
    \centering
    \includegraphics[width=\columnwidth]{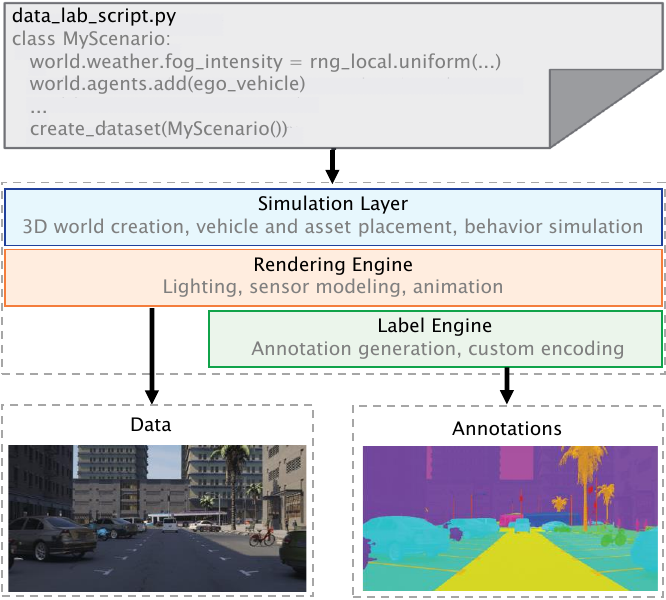}
    \caption{Overview of the \dsname{} generation pipeline: a user-controlled script, first automatically places and simulate agents in virtual 3D worlds, renders scenes under specified conditions and sensor arrangements, and finally generates ground truth annotations for a wide range of applications.}
    \label{fig:pipeline}
\vspace{-10px} \end{figure}

\section{Proposed Dataset : \dsname}

We propose \dsname{}, a high-fidelity all purpose multi-modal multi-task synthetic dataset for autonomous driving. Fig.~\ref{fig:pipeline} provides an overview of the generation pipeline, which can be broken down into three phases: simulation, rendering, and labeling. During simulation, agents are placed, and behavior is simulated based on the user's configuration. The rendering engine then renders various lighting conditions, visually realistic sensor outputs, and dynamic animations. Finally, the label engine generates ground-truth annotations required for training perception models. We generated \dsname{} using Parallel Domain's simulation environment\footnote{https://paralleldomain.com}. The scenarios were automatically configured using the Data Lab Python API for enhanced diversity. This auto-simulation includes placing agents, modeling their behavior, sampling weather and daytime conditions based on high-level scripting. 

\dsname{} contains 133,780 frames across 10 maps of urban, suburban, and highway environments. It provides 6343 normal condition sequences and 346 adverse condition sequences. Each sequence consists of 20 images sampled at 10Hz. These sequences span various times of the day: dawn, morning, noon, afternoon, dusk, and night. The adverse condition sequences contain night time, rainy, foggy, and mixed weather at various levels of severity. Specifically, there are 2,873 urban sequences, 1,497 suburban sequences, and 1,764 highway sequences in \dsname{}. 302 sequences with heavy fog and 70 sequences with heavy rain add to the diversity of \dsname{}. 92 dawn sequences, 82 dusk sequences, and 92 night sequences provide more realistic scenarios for autonomous driving. Fig.~\ref{fig:bigpdv4_example_samples} shows a few samples from our dataset in different weather and daytime conditions.

\begin{figure}[htbp]
    \centering
    \includegraphics[width=0.8\columnwidth]{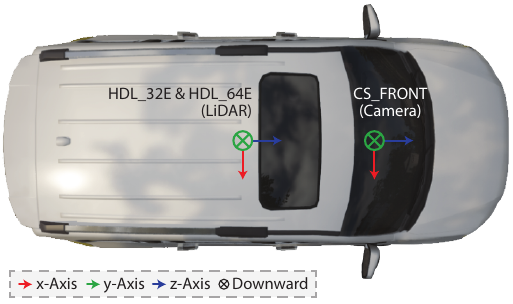}
    \caption{Illustration of the \dsname{} sensor configuration.}
    \label{fig:sensor_setup}
\vspace{-10px} \end{figure}

\subsection{Sensors}

We render \dsname{} using three sensors, each matching the intrinsics and extrinsics of a corresponding sensor from a popular public dataset:
\begin{enumerate}
    \item A front-facing RGB camera that matches Cityscapes in resolution and its intrinsic and extrinsic calibration.
    \item A 64-beam LiDAR sensor that matches the KITTI Velodyne HDL 64E.
    \item A 32-beam LiDAR sensor that matches the nuScenes Velodyne HDL 32E.
\end{enumerate}

All sensors operate at $10Hz$, allowing common preprocessing techniques of point cloud registration to be applied prior to training for datasets such as nuScenes, and also aiding research on real-time visual perception. An illustration of the sensor setup can be seen in Fig.~\ref{fig:sensor_setup}.

\subsection{Annotations}

In \dsname{}, we provide fine-grained annotations covering up to 64 class categories (Fig~\ref{fig:bigpd_class_stats}) for a wide range of tasks and supported modalities. The accompanying SDK\footnote{https://parallel-domain.github.io/pd-sdk/} provides not only class mappings to fully cover popular datasets but also individual dataset decoders to match the formatting of the target dataset for ease of use. In Tab.~\ref{tab:task_comp}, we summarise the currently supported tasks on \dsname{}.

\noindent\textbf{Object Detection:} We provide both 2D and 3D bounding box annotations for 53 object categories including but not limited to commonly seen traffic agents such as car, truck, bus, bicycle, motorcycle, and pedestrian. In addition to the 7 degrees of freedom commonly used in autonomous driving datasets, each bounding box is denoted by its roll, pitch, as well as its occlusion and truncation attributes. We further provide consistent instance IDs across each sequence to allow downstream tasks such as \textit{multi-object-tracking}.

\noindent\textbf{Panoptic Segmentation:} We provide pixel-level and point-level annotations covering up to 64 classes for both semantic and instance segmentation for 2D and 3D respectively. We support full class mappings and format conversions to many 2D datasets such as Cityscapes~\cite{Cordts2016Cityscapes} and BDD100k~\cite{yu2020bdd100k} and 3D datasets like SemanticKITTI~\cite{behley2019semantickitti} and SemanticPOSS~\cite{pan2020semanticposs}.

\noindent\textbf{Depth Estimation:} We provide pixel-level ground truth for depth estimation, denoting the distance between the pixel center and the sensor.

\noindent\textbf{Optical Flow:} We provide 2D motion vectors, mapping each pixel to the next frame.

\noindent\textbf{Scene Flow:} Similarly, we also provide 3D motion vectors that map the movement of each point between two frames relative to the ego vehicle motion.

\noindent\textbf{LiDAR Odometry and SLAM:} We provide the exact location and rotation of the ego vehicle in global coordinates for each frame allowing research into 3D odometry and SLAM. We also provide exact point-to-pixel correspondences.

\begin{table*}[htbp]
    \centering
    \tabcolsep=3.25pt
    \scriptsize
    \caption{\textbf{3D semantic segmentation} results on \textbf{SemanticKITTI}~\cite{behley2019semantickitti} using MinkowskiNet~\cite{choy20194d}. We compare synthetic LiDAR semantic segmentation datasets  in a source-only (SO) setting and when trained jointly with the real-world dataset.} 
    \begin{tabular}{|l|l|c|ccccccccccccccccccc|}
        \cline{2-22}
        \multicolumn{1}{c|}{} & Dataset 
        & mIoU
        &car
        &bicy.
        &motor.
        &truck 
        &o-veh.
        &pers.
        &b.list 
        &m.list 
        &road
        &park. 
        &s.walk
        &o-gro.
        &build.
        &fence
        &vege. 
        &trunk
        &terr.
        &pole
        &t-sign \\ 
        [0.5ex] 
        \cline{2-22}
        \multicolumn{1}{l|}{} & Oracle & 60.3 & 95.7 & 25.0 & 57.0 & 62.1 & 46.4 & 63.4 & 77.3 & 0.0 & 93.0 & 47.9 & 80.5 & 2.2 & 89.7 & 58.6 & 89.5 & 66.5 & 78.0 & 64.6 & 50.1 \\
        \hline
        \parbox[t]{3mm}{\multirow{2}{*}{\rotatebox[origin=c]{90}{SO}}} & SynLiDAR~\cite{xiao2022synlidar} & 20.4 & 42.0 & \textbf{5.0} & 4.8 & 0.4 & \textbf{2.5} & 12.4 & \textbf{43.3} & \textbf{1.8} & 48.7 & \textbf{4.5} & \textbf{31.0} & 0.0 & 18.6 & 11.5 & 60.2 & 30.0 & \textbf{48.3} & 19.3 & 3.0 \\
        & \dsname{} (Ours) & \textbf{28.2} & \textbf{82.8} & 1.1 & \textbf{21.3} & \textbf{3.3} & 0.0 & \textbf{20.3} & 38.5 & 1.4 & \textbf{65.0} & 4.1 & 16.5 & 0.0 & \textbf{61.4} & \textbf{36.1} & \textbf{76.7} & \textbf{41.3} & 23.7 & \textbf{31.4} & \textbf{11.0} \\
        \hline
        \parbox[t]{3mm}{\multirow{2}{*}{\rotatebox[origin=c]{90}{Joint}}} & SynLiDAR~\cite{xiao2022synlidar} & 62.5 & 95.9 & \textbf{33.0} & 62.8 & \textbf{78.9} & 50.2 & \textbf{71.4} & 83.5 & 0.7 & 92.3 & \textbf{52.8} & 79.9 & 0.1 & 89.8 & 59.5 & 86.3 & \textbf{65.4} & \textbf{72.8} & \textbf{63.6} & \textbf{48.9} \\
        & \dsname{} (Ours) & \textbf{63.1} & \textbf{97.2} & 26.2 & \textbf{65.8} & 76.5 & \textbf{66.4} & 69.6 & \textbf{84.2} & \textbf{1.0} & \textbf{93.1} & 52.1 & \textbf{80.6} & \textbf{2.6} & \textbf{91.0} & \textbf{63.3} & \textbf{86.5} & 62.1 & 70.8 & 62.8 & 47.1\\
        \hline
    \end{tabular}
    \label{tab:3dss_results_sk}
\vspace{-10px} \end{table*}

\begin{table*}[htbp]
    \tabcolsep=0.23cm
    \centering
    \footnotesize
    \caption{\textbf{3D semantic segmentation} results on \textbf{SemanticPOSS}~\cite{pan2020semanticposs} using MinkowskiNet~\cite{choy20194d}. We compare synthetic LiDAR semantic segmentation datasets  in a source-only (SO) setting and when trained jointly with the real-world dataset.} 
    \begin{tabular}{|l|l|c|ccccccccccccc|}
        \cline{2-16}
        \multicolumn{1}{l|}{} & Dataset
        & mIoU
        & pers.
        & rider
        & car
        & truck
        & plants
        & sign
        & pole
        & trash
        & build.
        & cone 
        & fence
        & bike
        & grou. \\
        [0.5ex] 
        \cline{2-16}
        \multicolumn{1}{l|}{} & Oracle & 50.6 & 55.6 & 45.1 & 66.9 & 44.4 & 73.9 & 45.4 & 41.6 & 14.5 & 76.1 & 7.9 & 57.0 & 54.1 & 75.3 \\
        \hline
        \parbox[t]{3mm}{\multirow{2}{*}{\rotatebox[origin=c]{90}{SO}}} & SynLiDAR~\cite{xiao2022synlidar} & 20.1 & 3.7 & 25.1 & 12.0 & 10.8 & \textbf{53.4} & 0.0 & \textbf{19.4} & 12.9 & \textbf{49.1} & 3.1 & 20.3 & 0.0 & \textbf{59.6} \\
        & \dsname{} (Ours) & \textbf{34.8} & \textbf{44.8} & \textbf{26.9} & \textbf{14.0} & \textbf{66.7} & 27.6 & \textbf{35.5} & 1.8 & \textbf{82.1} & 28.1 & \textbf{36.1} & \textbf{68.8} & \textbf{0.8} & 19.7 \\
        \hline
        \parbox[t]{3mm}{\multirow{2}{*}{\rotatebox[origin=c]{90}{Joint}}} & SynLiDAR~\cite{xiao2022synlidar} & 53.2 & 57.6 & \textbf{59.3} & 61.1 & 37.1 & \textbf{76.1} & 32.7 & 40.9 & 34.7 & \textbf{72.7} & 37.7 & 57.6 & 43.3 & \textbf{81.2} \\
        & \dsname{} (Ours) & \textbf{58.3} & \textbf{66.7} & 57.5 & \textbf{76.0} & \textbf{50.5} & 74.9 & \textbf{44.1} & \textbf{42.0} & \textbf{36.5} & 71.4 & \textbf{42.9} & \textbf{63.0} & \textbf{55.2} & 77.7 \\
        \hline
    \end{tabular}
    \label{tab:3dss_results_sp}
\vspace{-10px} \end{table*}

\begin{table}[htbp]
    \centering
    \footnotesize
    \tabcolsep=0.35cm
    \caption{\textbf{Fine-tuned 3D semantic segmentation} comparison on SemKITTI~\cite{behley2019semantickitti} and SemPOSS~\cite{pan2020semanticposs} using MinkowskiNet~\cite{choy20194d} Reported values are the mIoU $[\%]$ on the \textit{val}-set.} 
    \begin{tabular}{|l|cc|cc|}
         \cline{2-5}
         \multicolumn{1}{c|}{} & \multicolumn{2}{c|}{SemKITTI~\cite{behley2019semantickitti}} & \multicolumn{2}{c|}{SemPOSS~\cite{pan2020semanticposs}} \\
         \hline
         Labeled & SynL & \dsname{} & SynL & \dsname{} \\
         \hline
         $0\%$ & 20.4 & 28.2 & 20.1 & 34.8 \\
         $1\%$ & 40.4 & 49.8 & 36.2 & 44.1 \\
         $10\%$ & 54.0 & 56.1 & 46.0 & 50.0 \\
         $100\%$ & 61.9 & 63.5 & 53.2 & 59.0 \\
         \hline
    \end{tabular}
    \label{tab:3dss_finetuning}
\vspace{-10.1px} \end{table}
\section{Experiments}

In this section, we extensively evaluate \dsname{} across a wide range of popular tasks and datasets. Our focus is on assessing the raw quality of the datasets. Thus, to avoid any model or dataset-specific variability introduced by domain-adaptive methods, we only consider source-only, transfer learning, and joint training settings. Also, note that we remap our fine-grained class labels mentioned in Fig.~\ref{fig:bigpd_class_stats} to coarser labels as per the requirement of the task and the target dataset.

\subsection{3D Semantic Segmentation}

\noindent\textbf{Dataset:} We evaluate our synthetic dataset using the real-world datasets SemanticKITTI~\cite{behley2019semantickitti} and SemanticPOSS~\cite{pan2020semanticposs}.

\noindent\textbf{Implementation details:} For 3D semantic segmentation, we use the widely popular MinkowskiNet~\cite{choy20194d} in our experiments without modification. In the source-only or transfer learning setting, we only train on the source domain and evaluate on the real-world \textit{val}-set. For joint training, we use a balanced sampling scheme where each minibatch consists of an equal number of source and target frames. For fine-tuning, we reduce the learning rate by a factor of 10.

\noindent\textbf{Results:} In Tab.~\ref{tab:3dss_results_sk} we compare the performance of our dataset to SynLiDAR~\cite{xiao2022synlidar} in a source-only and a joint training setting when evaluated on SemanticKITTI. As seen, \dsname{} provides significant improvements over SynLiDAR on both tasks, boasting up to a $+2.8\%$ mIoU improvement over the oracle (trained and evaluated on the source dataset) MinkowskiNet. In Tab.~\ref{tab:3dss_results_sp} we show a similar study on the SemanticPOSS dataset. As with the former study on SemanticKITTI, \dsname{} outperforms SynLiDAR significantly, showing gains of $+7.7\%$ mIoU over the oracle when trained jointly with the real-world dataset.

Furthermore, in Tab.~\ref{tab:3dss_finetuning} we demonstrate the usefulness of \dsname{} under fine-tuning conditions. Pretrained on the synthetic dataset, we fine-tune our model on subsets of the real-world dataset at varying sampling ratios. The results show that pretraining with \dsname{} can drastically reduce the gap to the oracle performance. Furthermore, pretraining on \dsname{} outperforms pretraining on SynLiDAR across the board.

\subsection{3D Object Detection}

\begin{table*}[htbp]
    \centering
    \footnotesize
    \tabcolsep=0.25cm
    \caption{\textbf{3D object detection} results on the KITTI~\cite{geiger2012we} \textit{val}-set using SECOND~\cite{yan2018second} reporting mAP@0.7 for the ``Car" category under BEV, 3D box and orientation [\%]. We show both a source-only (SO) setting and when trained jointly with KITTI.} 
    \begin{tabular}{|l|l|ccc|ccc|ccc|}
    \cline{3-11}
    \multicolumn{2}{c|}{} & \multicolumn{3}{c|}{BEV} & \multicolumn{3}{c|}{3D} & \multicolumn{3}{c|}{AOS} \\
    \cline{2-11}
    \multicolumn{1}{c|}{} & Dataset & Easy & Moderate & Hard & Easy & Moderate & Hard & Easy & Moderate & Hard \\
    \cline{2-11}
    \multicolumn{1}{c|}{} & Oracle & 88.92 & 85.67 & 84.18 & 86.55 & 76.20 & 73.31 & 90.59 & 89.30 & 88.45 \\
    \hline
        \parbox[t]{2mm}{\multirow{2}{*}{\rotatebox[origin=c]{90}{SO}}} & PreSIL~\cite{hurl2019precise_presil} & 26.49 & 27.23 & 28.28 & 12.27 & 12.21 & 12.55  & 71.55 & 61.33 & 62.22 \\
        & \dsname{} (Ours) & \textbf{34.08} & \textbf{33.27} & \textbf{34.19} & \textbf{16.32} & \textbf{16.52} & \textbf{15.94} & \textbf{78.16} & \textbf{67.65} & \textbf{69.03} \\
        \hline
        \parbox[t]{2mm}{\multirow{2}{*}{\rotatebox[origin=c]{90}{Joint}}} & PreSIL~\cite{hurl2019precise_presil} & \textbf{89.51} & 84.25 & 84.41 & 86.33 & 75.17 & 72.81 & 90.67 & 89.08 & 87.90 \\
        & \dsname{} (Ours) & 89.50 & \textbf{86.41} & \textbf{84.79} & \textbf{87.47} & \textbf{76.76} & \textbf{74.95} & \textbf{95.95} & \textbf{89.47} & \textbf{88.43} \\
    \hline
    \end{tabular}
    \label{tab:3dod_results}
\vspace{-10px} \end{table*}

\noindent\textbf{Dataset:} We evaluate our method on the KITTI~\cite{geiger2012we} dataset for the popular ``Car" category.

\noindent\textbf{Implementation details:} In our experiments for 3D object detection, we use the OpenPCDet~\cite{openpcdet2020} implementation of SECOND~\cite{yan2018second} without the copy-paste augmentation. Similar to 3D semantic segmentation, for joint training, we use a balanced sampling scheme where each minibatch consists of equal source and target frames.

\noindent\textbf{Results:} In Tab.~\ref{tab:3dod_results} we demonstrate the effectiveness of \dsname{} on source-only and joint training. As seen, \dsname{} consistently outperforms the competing PreSIL~\cite{hurl2019precise_presil} on bird's-eye-view (BEV) detection, 3D bounding box estimation and on average orientation similarity (AOS). Specifically, for 3D object detection, training with \dsname{} shows $+4.31\%$ and $+1.59\%$ mAP improvements for the moderate category in source-only and joint training settings respectively.

\subsection{2D Object Detection}
\label{exp:2d_od}
 
\begin{table}[htbp]
	\centering
	\footnotesize
        \tabcolsep=0.075cm
    \caption{\textbf{Source-only 2D object detection} AP with 0.5 IoU threshold in \% on Cityscapes and BDD100k \textit{val}-SET. The models are trained on different synthetic datasets.
    The best performance is highlighted with bold text.} 
    \begin{tabular}{|l|l|cccccccc|c|}
    \cline{2-11}
    \multicolumn{1}{c|}{} & Training          & \rotatebox{90}{Person} & \rotatebox{90}{Rider}  & \rotatebox{90}{Car}   & \rotatebox{90}{Truck}  & \rotatebox{90}{Bus} & \rotatebox{90}{Train} & \rotatebox{90}{M.cycle} & \rotatebox{90}{B.cycle} & \rotatebox{90}{mAP} \\
    \hline
    \parbox[t]{2.5mm}{\multirow{4}{*}{\rotatebox[origin=c]{90}{Cityscapes}}} & SHIFT \cite{sun2022shift}     & 39.8  & 0.0      & 44.2  & 5.7    & 3.3  & 0.0    & 13.5   & 12.4  & 14.9 \\
    & VIPER \cite{Richter_2017_ICCV_Viper}     & 27.7  & 0.0      & 29.8  &  9.6  & 5.9  & 0.0    & 6.1    & 0.0     & 9.9 \\
    & Synscapes \cite{wrenninge2018synscapes} & \textbf{61.6}  & 59.3   & \textbf{74.3}  & \textbf{17.9}   & 49.2 & 34.7 & 22.3   & 33.3  & 44.1\\
    & \dsname{} (Ours) \       & 56.4 & \textbf{67.4} & 67.6 & 17.7 & \textbf{50.9} & \textbf{47.5} & \textbf{32.5} & \textbf{49.0} & \textbf{48.6} \\
    \hline
    \parbox[t]{2.5mm}{\multirow{4}{*}{\rotatebox[origin=c]{90}{BDD100k}}} & SHIFT \cite{sun2022shift}     & 31.0  & 0.0      & 26.0  & 3.2    &  0.6 & 0.0    & 8.1    & 9.3   & 9.8 \\
    & VIPER \cite{Richter_2017_ICCV_Viper}     & 17.3  & 0.0      & 26.8  & 6.2    & 4.4  & 0.0    & 3.2    & 0.0     & 7.2 \\
    & Synscapes \cite{wrenninge2018synscapes} & \textbf{31.6}  &  \textbf{22.8}  & 39.5  & 7.7    & 8.9  & 0.0  & 8.1    & 14.5  & 16.7 \\
    & \dsname{} (Ours) \       &  27.8 & 19.4 & \textbf{44.9} & \textbf{13.9} & \textbf{24.8} &  0.0 & \textbf{11.4} & \textbf{18.9} & \textbf{20.1}\\
    \hline
	\end{tabular}
    \label{tab:2dod_so_res}
\vspace{-10px} \end{table}

\begin{table}[htbp]
	\centering
	\footnotesize
        \tabcolsep=0.08cm
    \caption{\textbf{Fine-tuned 2D object detection} AP with 0.5 IoU threshold in \% on Cityscapes and BDD100k \textit{val}-SET. The models are pre-trained on different synthetic datasets and finetuned on Cityscapes and BDD \textit{train}-set. The best performance is highlighted with bold text.} 
    \begin{tabular}{|l|l|cccccccc|c|}
    \cline{2-11}
    \multicolumn{1}{c|}{} &Pretrained Data   & \rotatebox{90}{Person} & \rotatebox{90}{Rider}  & \rotatebox{90}{Car}   & \rotatebox{90}{Truck}  & \rotatebox{90}{Bus} & \rotatebox{90}{Train} & \rotatebox{90}{M.cycle} & \rotatebox{90}{B.cycle} & \rotatebox{90}{mAP} \\
    \hline
    \parbox[t]{2.5mm}{\multirow{4}{*}{\rotatebox[origin=c]{90}{Cityscapes}}} & SHIFT \cite{sun2022shift}     &  66.3 & \textbf{69.6}   & 79.0  & 47.6   & 72.7   & 59.3    &53.4 & 56.3 & 63.0     \\
    &VIPER \cite{Richter_2017_ICCV_Viper}     & 66.3  & 66.1   & 78.5  & 45.4   & 69.9  & 62.8 & 49.9 & 54.8 &61.7 \\
    &Synscapes \cite{wrenninge2018synscapes} & \textbf{68.6}  & 68.3   & \textbf{79.8}  & 48.9   & 72.2  & 52.4 & 53.3 & 57.4 & 62.6   \\
    &\dsname{} (Ours)       & 68.4 & 68.7 & 78.8 & \textbf{49.0} & \textbf{74.5} & \textbf{65.4} & \textbf{54.4} & \textbf{58.9} & \textbf{64.8}\\
    \hline
    \parbox[t]{2.5mm}{\multirow{4}{*}{\rotatebox[origin=c]{90}{BDD100k}}} &SHIFT \cite{sun2022shift}     &  \textbf{38.2} & \textbf{26.8}   & 41.6  & 15.9   & 13.0   & 0.0    &12.7 & \textbf{19.1} &  20.9    \\
    &VIPER \cite{Richter_2017_ICCV_Viper}     & 35.5  & 25.3   & 42.9  &  17.9  & 13.4  & 0.0 & 11.5 &16.0& 20.3 \\
    &Synscapes \cite{wrenninge2018synscapes} & 36.7  &  25.5  & \textbf{43.0}  & 14.8   & 13.7  & 0.0 & 12.7 & 18.4 &  20.6  \\
    &\dsname{} (Ours)       &  36.8 & 25.8 & 42.2 & \textbf{16.3} &  \textbf{16.5} &0.0 & \textbf{14.4} & 16.5 & \textbf{21.1}\\
    \hline
    \end{tabular}
    \label{tab:2dod_ft_res}
\vspace{-10px} \end{table}

\noindent\textbf{Dataset:} For 2D object detection, we compare our dataset with existing synthetic datasets including VIPER \cite{Richter_2017_ICCV_Viper}, Synscapes \cite{wrenninge2018synscapes}, and SHIFT \cite{sun2022shift}. %
For the evaluation of trained models, we consider the validation set of two real-world datasets. A clear weather European dataset, Cityscapes~\cite{Cordts2016Cityscapes} and an American mixed weather dataset, BDD100k~\cite{yu2020bdd100k}.

\noindent\textbf{Implementation details:} Our evaluation is based on the MMDetection framework \cite{mmdetection}. For a fair comparison, we choose the classic object detector Faster R-CNN \cite{ren2015faster_rcnn} for each experiment. Since the size of SHIFT and \dsname{} is much larger than other synthetic datasets, following set precedent~\cite{sun2022shift}, we uniformly sample frames from each sequence to subsample a dataset of the same size as other competing datasets. We therefore only train with 32,292 images uniformly sampled from \dsname{}. We compare different synthetic datasets in the source-only training and fine-tuning settings.

\noindent\textbf{Results:} The evaluation results of our source-only training are presented in Tab.~\ref{tab:2dod_so_res}. We observe that \dsname{} presents a better generalization ability than other synthetic datasets. On both Cityscapes and BDD100k validation sets, the model trained on \dsname{} brings over $+3\%$ improvement compared to the second-best model trained on Synscapes.

Tab.~\ref{tab:2dod_ft_res} demonstrates the fine-tuning performance with different pre-trained models. Compared to other synthetic datasets, \dsname{} dataset pretraining brings the largest mAP improvement on both Cityscapes and BDD100k.

\subsection{2D Semantic Segmentation} \label{sec_2d_seg}

\begin{table*}[htbp]
	\centering
	\footnotesize
	\setlength{\tabcolsep}{3.5pt}
     \caption{\textbf{Source-only 2D semantic segmentation} results on Cityscapes~\cite{Cordts2016Cityscapes} and BDD100k~\cite{yu2020bdd100k} \textit{val}-set. The results
of each model are averaged over 3 random seeds The best performance is highlighted with bold text and the second best is underlined. 
    *Synscapes was built to match the Cityscapes distribution, while ours is not tuned towards a specific target domain.} 
    \begin{tabular}{|l|l|ccccccccccccccccccc|c|}
    \cline{2-22}
    \multicolumn{1}{c|}{} & Dataset & \rotatebox{90}{Road} & \rotatebox{90}{S.walk} & \rotatebox{90}{Build.} & \rotatebox{90}{Wall} & \rotatebox{90}{Fence} & \rotatebox{90}{Pole} & \rotatebox{90}{Tr. Light \hphantom{o}} & \rotatebox{90}{Tr. Sign} & \rotatebox{90}{Veget.} & \rotatebox{90}{Terrain} & \rotatebox{90}{Sky} & \rotatebox{90}{Person} & \rotatebox{90}{Rider} & \rotatebox{90}{Car} & \rotatebox{90}{Truck} & \rotatebox{90}{Bus} & \rotatebox{90}{Train} & \rotatebox{90}{M.cycle} & \rotatebox{90}{Bicycle} & \rotatebox{90}{mIoU} \\
    \hline
    \parbox[t]{2.5mm}{\multirow{5}{*}{\rotatebox[origin=c]{90}{Cityscapes}}} & GTA \cite{richter2016playing_GTA} & 74.5 & 17.9 & \underline{84.5} & \underline{29.0} & \underline{31.8} & 30.5 & \underline{43.2} & 19.1 & \textbf{86.9} & \underline{40.2} & \textbf{87.2} & \underline{62.3} & 20.9 & \underline{87.8} & \textbf{44.1} & \underline{42.4} & 1.5 & \textbf{32.1} & 21.8 & 45.1 \\
    & Synscapes* \cite{wrenninge2018synscapes} & 91.1 & 46.1 & 81.2 & \textbf{33.8} &\textbf{ 37.4} & \textbf{41.7} & \textbf{43.6} & \textbf{49.8} & \underline{85.9} & \textbf{46.1} & 85.4 & \textbf{66.1} & \textbf{35.6} & 87.3 & 23.3 & 26.7 & \underline{13.1} & \underline{28.6} & \textbf{59.6} & \textbf{51.7} \\
    & Synthia \cite{ros2016synthia_synthia_rand_seqs} & 41.7 & 18.6 & 80.5 & 19.0 & 1.5 & \underline{38.8} & 32.2 & 19.4 & 82.7 & 0.0 & 78.0 & 61.3 & 19.8 & 72.2 & 0.0 & 33.5 & 0.0 & 14.7 & 16.7 & 33.2 \\
    & SHIFT \cite{sun2022shift} & \textbf{93.9} & \textbf{60.9} & 82.1 & 10.1 & 3.4 & 37.4 & 21.6 & 31.8 & 85.6 & 30.7 & 86.4 & 54.8 & 0.0 & 77.6 & 0.0 & 0.0 & 0.0 & 0.0 & 0.0 & 35.6 \\
    & \dsname{} (Ours) & \underline{92.7} & \underline{54.3} & \textbf{85.1} & 16.8 & 5.9 & 33.3 & 31.1 & \underline{32.8} & 84.8 & 18.7 & \underline{86.5} & 53.1 & \underline{29.9} & \textbf{87.9} & \underline{41.7} & \textbf{44.8} & \textbf{19.2} & 27.1 & \underline{52.8} & \underline{47.4} \\
    \hline
    \parbox[t]{2.5mm}{\multirow{5}{*}{\rotatebox[origin=c]{90}{BDD100k}}} & GTA \cite{richter2016playing_GTA} & 83.4 & 20.8 & \textbf{77.1} & \textbf{21.4} & \textbf{36.5} & \textbf{33.1} & \textbf{38.4} & 26.0 & \textbf{74.3} & \textbf{35.9} & 88.2 & \underline{54.2} & \textbf{25.7} & \textbf{80.0} & \textbf{29.0} & \underline{35.8} & 0.0 & \textbf{43.2} & \underline{27.0} & \textbf{43.7}\\
    & Synscapes* \cite{wrenninge2018synscapes} & \underline{85.7} & 28.0 & 61.4 & 7.1 & \underline{19.0} & 26.4 & 26.1 & 24.3 & 71.9 & 21.0 & 87.6 & 43.2 & \underline{19.0} & \underline{79.6} & 8.5 & 16.8 & 0.0 & 21.6 & 25.4 & 35.4 \\
    & Synthia \cite{ros2016synthia_synthia_rand_seqs} & 21.7 & 10.7 & 68.3 & 3.5 & 0.0 & 28.3 & \underline{37.7} & 23.6 & 69.3 & 0.0 & 87.7 & 44.8 & 14.5 & 56.9 & 0.0 & 31.2 & 0.0 & 12.4 & 18.7 & 27.9\\
    & SHIFT \cite{sun2022shift} & \textbf{88.1} & \textbf{39.2} & \underline{76.5} & \underline{21.1} & 7.0 & \underline{31.9} & 37.1 & \textbf{35.7} & \underline{73.3} & 25.1 & \textbf{90.6} & \textbf{54.8} & 0.0 & 75.7 & 0.0 & 0.0 & 0.0 & 0.0 & 0.0 & 34.5 \\
    & \dsname{} (Ours) & 84.0 & \underline{35.6} & 73.8 & 11.3 & 8.9 & 29.8 & 29.8 & \underline{29.3} & 71.8 & \underline{27.7} & \underline{89.2} & 27.1 & 5.4 & 78.8 & \underline{27.5} & \textbf{41.0} & 0.0 & \underline{22.0} & \textbf{42.0} & \underline{38.7} \\
    \hline
    \end{tabular}
    \label{tab:2dsseg_so}
\vspace{-10px} \end{table*}

\noindent\textbf{Dataset:} For 2D semantic segmentation, we consider four commonly followed synthetic datasets for training: GTA~\cite{richter2016playing_GTA}, Synscapes~\cite{wrenninge2018synscapes}, SHIFT~\cite{sun2022shift}, and Synthia~\cite{ros2016synthia_synthia_rand_seqs}. Same as section~\ref{exp:2d_od} we use Cityscales and BDD100k for evaluating the trained models. Following prior works~\cite{hoyer2022daformer, xie2021segformer}, we conduct evaluations using the 19 Cityscapes evaluation classes. 

\noindent\textbf{Implementation details:}
We compare the source-only performance of different synthetic datasets using the simple and commonly followed Segformer~\cite{xie2021segformer} architecture with a MiT-B5 encoder. 
Following set precedent~\cite{hoyer2022daformer}, we downsample the images by a factor of two, train with 512x512 crops, and a batch size of 2 for fast and memory-efficient training. We follow the same dataset subsampling strategy as in the 2D object detection experiments.

\noindent\textbf{Results:}
Tab.~\ref{tab:2dsseg_so} shows the semantic segmentation performance of different synthetic datasets on Cityscapes and BDD100k.
As stated earlier, we employ a source-only training setup to evaluate the synthetic-to-real transfer performance of raw synthetic datasets without introducing any external techniques, which could influence the results differently depending on the source dataset. As can be seen, \dsname{} comfortably outperforms GTA, SYNTHIA, and SHIFT on Cityscapes. However, Synscapes achieves better performance than \dsname{}, which is expected since it is specifically designed to match Cityscapes in both structure and content. This is further verified with our experiments on BDD100k. Since BDD is captured in an American environment, it is significantly different than Cityscapes from the structure and content perspective. Consequently, Synscapes' performance drops by a huge margin for BDD100k. \dsname{} on the other hand, is the second-best performing dataset.
Again, \dsname{} is only outperformed by GTA, which simulates US scenes, similar to BDD.
Overall, \dsname{} achieves a better balance between these target domains as it is not tailored to a specific one.
\dsname{} consistently performs well for vehicle classes across different target real datasets. Moreover, the lack of multiple semantic classes puts SHIFT and Synthia at a substantial disadvantage compared to other datasets in practical applications.

\vspace{-1.0px}

\section{Conclusion}

We generate and introduce \dsname{}, a high-fidelity  multi-modal multi-task synthetic dataset for autonomous driving. \dsname{} not only supports popular camera-based 2D perception tasks (e.g., 2D semantic segmentation, 2D object detection, depth estimation, optical flow) but also provides fine-grained annotations for a wide range of LiDAR-based 3D perception tasks (e.g. 3D object detection, 3D semantic segmentation, and Scene Flow). We show extensive studies on \dsname{} by testing on various real-world datasets and demonstrate its advantage on tasks such as source-only training, joint training, and fine-tuning when compared to existing synthetic datasets.

{\small
\bibliographystyle{IEEEtran}

\begin{thebibliography}{10}
\providecommand{\url}[1]{#1}
\csname url@samestyle\endcsname
\providecommand{\newblock}{\relax}
\providecommand{\bibinfo}[2]{#2}
\providecommand{\BIBentrySTDinterwordspacing}{\spaceskip=0pt\relax}
\providecommand{\BIBentryALTinterwordstretchfactor}{4}
\providecommand{\BIBentryALTinterwordspacing}{\spaceskip=\fontdimen2\font plus
\BIBentryALTinterwordstretchfactor\fontdimen3\font minus \fontdimen4\font\relax}
\providecommand{\BIBforeignlanguage}[2]{{%
\expandafter\ifx\csname l@#1\endcsname\relax
\typeout{** WARNING: IEEEtran.bst: No hyphenation pattern has been}%
\typeout{** loaded for the language `#1'. Using the pattern for}%
\typeout{** the default language instead.}%
\else
\language=\csname l@#1\endcsname
\fi
#2}}
\providecommand{\BIBdecl}{\relax}
\BIBdecl

\bibitem{zhang2021survey}
Y.~Zhang and Q.~Yang, ``A survey on multi-task learning,'' \emph{IEEE Transactions on Knowledge and Data Engineering}, vol.~34, no.~12, 2021.

\bibitem{unal2021improving}
O.~Unal, L.~Van~Gool, and D.~Dai, ``Improving point cloud semantic segmentation by learning 3d object detection,'' in \emph{WACV}, 2021.

\bibitem{bhattacharjee2022mult}
D.~Bhattacharjee, T.~Zhang, S.~S{\"u}sstrunk, and M.~Salzmann, ``Mult: An end-to-end multitask learning transformer,'' in \emph{CVPR}, 2022.

\bibitem{li2022deepfusion}
Y.~Li, A.~W. Yu, T.~Meng, B.~Caine, J.~Ngiam, D.~Peng, J.~Shen, Y.~Lu, D.~Zhou, Q.~V. Le \emph{et~al.}, ``Deepfusion: Lidar-camera deep fusion for multi-modal 3d object detection,'' in \emph{CVPR}, 2022.

\bibitem{unal2022scribble}
O.~Unal, D.~Dai, and L.~Van~Gool, ``Scribble-supervised lidar semantic segmentation,'' in \emph{CVPR}, 2022, pp. 2697--2707.

\bibitem{xu2020weakly}
X.~Xu and G.~H. Lee, ``Weakly supervised semantic point cloud segmentation: Towards 10x fewer labels,'' in \emph{CVPR}, 2020.

\bibitem{zhang2021weakly}
Y.~Zhang, Z.~Li, Y.~Xie, Y.~Qu, C.~Li, and T.~Mei, ``Weakly supervised semantic segmentation for large-scale point cloud,'' in \emph{AAAI}, 2021.

\bibitem{unal2023discwise}
O.~Unal, D.~Dai, A.~T. Unal, and L.~Van~Gool, ``Discwise active learning for lidar semantic segmentation,'' \emph{IEEE RA-L}, 2023.

\bibitem{wang2022semi}
Y.~Wang, H.~Wang, Y.~Shen, J.~Fei, W.~Li, G.~Jin, L.~Wu, R.~Zhao, and X.~Le, ``Semi-supervised semantic segmentation using unreliable pseudo-labels,'' in \emph{CVPR}, 2022, pp. 4248--4257.

\bibitem{kwon2022semi}
D.~Kwon and S.~Kwak, ``Semi-supervised semantic segmentation with error localization network,'' in \emph{CVPR}, 2022, pp. 9957--9967.

\bibitem{richter2016playing_GTA}
S.~R. Richter, V.~Vineet, S.~Roth, and V.~Koltun, ``Playing for data: Ground truth from computer games,'' in \emph{ECCV}.\hskip 1em plus 0.5em minus 0.4em\relax Springer, 2016.

\bibitem{Richter_2017_ICCV_Viper}
S.~R. Richter, Z.~Hayder, and V.~Koltun, ``Playing for benchmarks,'' in \emph{ICCV}, Oct 2017.

\bibitem{hurl2019precise_presil}
B.~Hurl, K.~Czarnecki, and S.~Waslander, ``Precise synthetic image and lidar (presil) dataset for autonomous vehicle perception,'' in \emph{IV}.\hskip 1em plus 0.5em minus 0.4em\relax IEEE, 2019, pp. 2522--2529.

\bibitem{Dosovitskiy17_CARLA_Sim}
A.~Dosovitskiy, G.~Ros, F.~Codevilla, A.~Lopez, and V.~Koltun, ``{CARLA}: {An} open urban driving simulator,'' in \emph{CoRL}, 2017, pp. 1--16.

\bibitem{sun2022shift}
T.~Sun, M.~Segu, J.~Postels, Y.~Wang, L.~Van~Gool, B.~Schiele, F.~Tombari, and F.~Yu, ``Shift: a synthetic driving dataset for continuous multi-task domain adaptation,'' in \emph{CVPR}, 2022, pp. 21\,371--21\,382.

\bibitem{alberti2020idda}
E.~Alberti, A.~Tavera, C.~Masone, and B.~Caputo, ``Idda: A large-scale multi-domain dataset for autonomous driving,'' \emph{IEEE Robotics and Automation Letters}, vol.~5, no.~4, pp. 5526--5533, 2020.

\bibitem{sekkat2022synwoodscape}
A.~R. Sekkat, Y.~Dupuis, V.~R. Kumar, H.~Rashed, S.~Yogamani, P.~Vasseur, and P.~Honeine, ``Synwoodscape: Synthetic surround-view fisheye camera dataset for autonomous driving,'' \emph{IEEE Robotics and Automation Letters}, vol.~7, no.~3, pp. 8502--8509, 2022.

\bibitem{sekkat2024amodalsynthdrive}
A.~R. Sekkat, R.~Mohan, O.~Sawade, E.~Matthes, and A.~Valada, ``Amodalsynthdrive: A synthetic amodal perception dataset for autonomous driving,'' \emph{IEEE Robotics and Automation Letters}, 2024.

\bibitem{Cordts2016Cityscapes}
M.~Cordts, M.~Omran, S.~Ramos, T.~Rehfeld, M.~Enzweiler, R.~Benenson, U.~Franke, S.~Roth, and B.~Schiele, ``The cityscapes dataset for semantic urban scene understanding,'' in \emph{CVPR}, 2016.

\bibitem{varma2019idd}
G.~Varma, A.~Subramanian, A.~Namboodiri, M.~Chandraker, and C.~Jawahar, ``Idd: A dataset for exploring problems of autonomous navigation in unconstrained environments,'' in \emph{2019 IEEE winter conference on applications of computer vision (WACV)}.\hskip 1em plus 0.5em minus 0.4em\relax IEEE, 2019, pp. 1743--1751.

\bibitem{yu2020bdd100k}
F.~Yu, H.~Chen, X.~Wang, W.~Xian, Y.~Chen, F.~Liu, V.~Madhavan, and T.~Darrell, ``Bdd100k: A diverse driving dataset for heterogeneous multitask learning,'' in \emph{CVPR}, 2020, pp. 2636--2645.

\bibitem{sakaridis2024acdcadverseconditionsdataset}
\BIBentryALTinterwordspacing
C.~Sakaridis, H.~Wang, K.~Li, R.~Zurbrügg, A.~Jadon, W.~Abbeloos, D.~O. Reino, L.~V. Gool, and D.~Dai, ``Acdc: The adverse conditions dataset with correspondences for robust semantic driving scene perception,'' 2024. [Online]. Available: \url{https://arxiv.org/abs/2104.13395}
\BIBentrySTDinterwordspacing

\bibitem{vu2019advent}
T.-H. Vu, H.~Jain, M.~Bucher, M.~Cord, and P.~P{\'e}rez, ``Advent: Adversarial entropy minimization for domain adaptation in semantic segmentation,'' in \emph{Proceedings of the IEEE/CVF conference on computer vision and pattern recognition}, 2019, pp. 2517--2526.

\bibitem{hoyer2022daformer}
L.~Hoyer, D.~Dai, and L.~Van~Gool, ``Daformer: Improving network architectures and training strategies for domain-adaptive semantic segmentation,'' in \emph{CVPR}, 2022.

\bibitem{loiseau2024reliability}
T.~Loiseau, T.-H. Vu, M.~Chen, P.~P{\'e}rez, and M.~Cord, ``Reliability in semantic segmentation: Can we use synthetic data?'' in \emph{ECCV}, 2024.

\bibitem{ros2016synthia_synthia_rand_seqs}
G.~Ros, L.~Sellart, J.~Materzynska, D.~Vazquez, and A.~M. Lopez, ``The synthia dataset: A large collection of synthetic images for semantic segmentation of urban scenes,'' in \emph{CVPR}, 2016, pp. 3234--3243.

\bibitem{xiao2022synlidar}
A.~Xiao, J.~Huang, D.~Guan, F.~Zhan, and S.~Lu, ``Transfer learning from synthetic to real lidar point cloud for semantic segmentation,'' in \emph{AAAI}, vol.~36, no.~3, 2022, pp. 2795--2803.

\bibitem{behley2019semantickitti}
J.~Behley, M.~Garbade, A.~Milioto, J.~Quenzel, S.~Behnke, C.~Stachniss, and J.~Gall, ``Semantickitti: A dataset for semantic scene understanding of lidar sequences,'' in \emph{ICCV}, 2019, pp. 9297--9307.

\bibitem{pan2020semanticposs}
Y.~Pan, B.~Gao, J.~Mei, S.~Geng, C.~Li, and H.~Zhao, ``Semanticposs: A point cloud dataset with large quantity of dynamic instances,'' in \emph{IV}.\hskip 1em plus 0.5em minus 0.4em\relax IEEE, 2020, pp. 687--693.

\bibitem{geiger2012we}
A.~Geiger, P.~Lenz, and R.~Urtasun, ``Are we ready for autonomous driving? the kitti vision benchmark suite,'' in \emph{CVPR}.\hskip 1em plus 0.5em minus 0.4em\relax IEEE, 2012, pp. 3354--3361.

\bibitem{wrenninge2018synscapes}
M.~Wrenninge and J.~Unger, ``Synscapes: A photorealistic synthetic dataset for street scene parsing,'' \emph{arXiv preprint arXiv:1810.08705}, 2018.

\bibitem{choy20194d}
C.~Choy, J.~Gwak, and S.~Savarese, ``4d spatio-temporal convnets: Minkowski convolutional neural networks,'' in \emph{CVPR}, 2019, pp. 3075--3084.

\bibitem{yan2018second}
Y.~Yan, Y.~Mao, and B.~Li, ``Second: Sparsely embedded convolutional detection,'' \emph{Sensors}, vol.~18, no.~10, p. 3337, 2018.

\bibitem{openpcdet2020}
O.~D. Team, ``Openpcdet: An open-source toolbox for 3d object detection from point clouds,'' \url{https://github.com/open-mmlab/OpenPCDet}, 2020.

\bibitem{mmdetection}
K.~Chen, J.~Wang, J.~Pang, Y.~Cao, Y.~Xiong, X.~Li, S.~Sun, W.~Feng, Z.~Liu, J.~Xu, Z.~Zhang, D.~Cheng, C.~Zhu, T.~Cheng, Q.~Zhao, B.~Li, X.~Lu, R.~Zhu, Y.~Wu, J.~Dai, J.~Wang, J.~Shi, W.~Ouyang, C.~C. Loy, and D.~Lin, ``{MMDetection}: Open mmlab detection toolbox and benchmark,'' \emph{arXiv preprint arXiv:1906.07155}, 2019.

\bibitem{ren2015faster_rcnn}
S.~Ren, K.~He, R.~Girshick, and J.~Sun, ``Faster r-cnn: Towards real-time object detection with region proposal networks,'' in \emph{NIPS}.\hskip 1em plus 0.5em minus 0.4em\relax Cambridge, MA, USA: MIT Press, 2015.

\bibitem{xie2021segformer}
E.~Xie, W.~Wang, Z.~Yu, A.~Anandkumar, J.~M. Alvarez, and P.~Luo, ``Segformer: Simple and efficient design for semantic segmentation with transformers,'' \emph{NIPS}, vol.~34, pp. 12\,077--12\,090, 2021.

\end{thebibliography}

}

\end{document}